\documentclass[10pt,twocolumn,letterpaper]{article}
\usepackage{graphicx}
\usepackage{iccv}      % To produce the REVIEW 
\usepackage{xcolor}
\usepackage{color,colortbl}
\definecolor{Gray2}{gray}{0.95}
\usepackage{pifont}
\usepackage{multirow} 
\usepackage{makecell}
\usepackage{colortbl}
\usepackage{longtable}

\definecolor{iccvblue}{rgb}{0.21,0.49,0.74}
\usepackage[pagebackref,breaklinks,colorlinks,allcolors=iccvblue]{hyperref}

\title{\includegraphics[height=0.7cm]{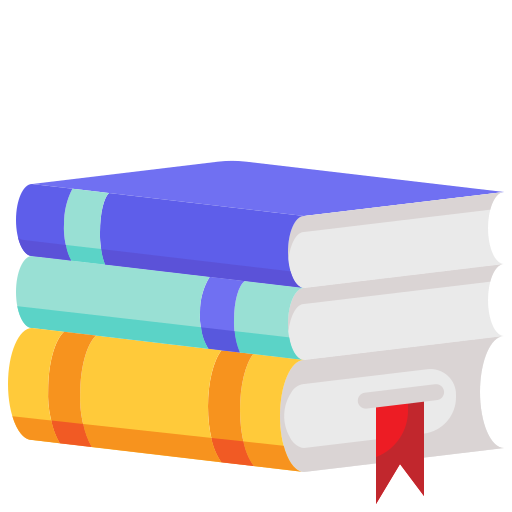} \hspace{0cm} Derm1M: A Million-scale Vision-Language Dataset Aligned with Clinical Ontology Knowledge for Dermatology}
\author{
Siyuan Yan$^{1}$\thanks{Equal contribution}\quad
Ming Hu$^{1*}$\quad
Yiwen Jiang$^{1*}$ \quad
Xieji Li$^{1}$\quad\\
Hao Fei$^{2}$\quad
Philipp Tschandl$^3$\quad
Harald Kittler
$^3$\quad
Zongyuan Ge$^{1}$\quad\\
$^1$ Monash University\quad
$^2$ National University of Singapore\quad
$^3$ Medical University of Vienna \quad\\
}

\makeatletter
\let\@oldmaketitle\@maketitle
\renewcommand{\@maketitle}{\@oldmaketitle
\vspace{-5pt}
\centering
\includegraphics[width=1\linewidth]{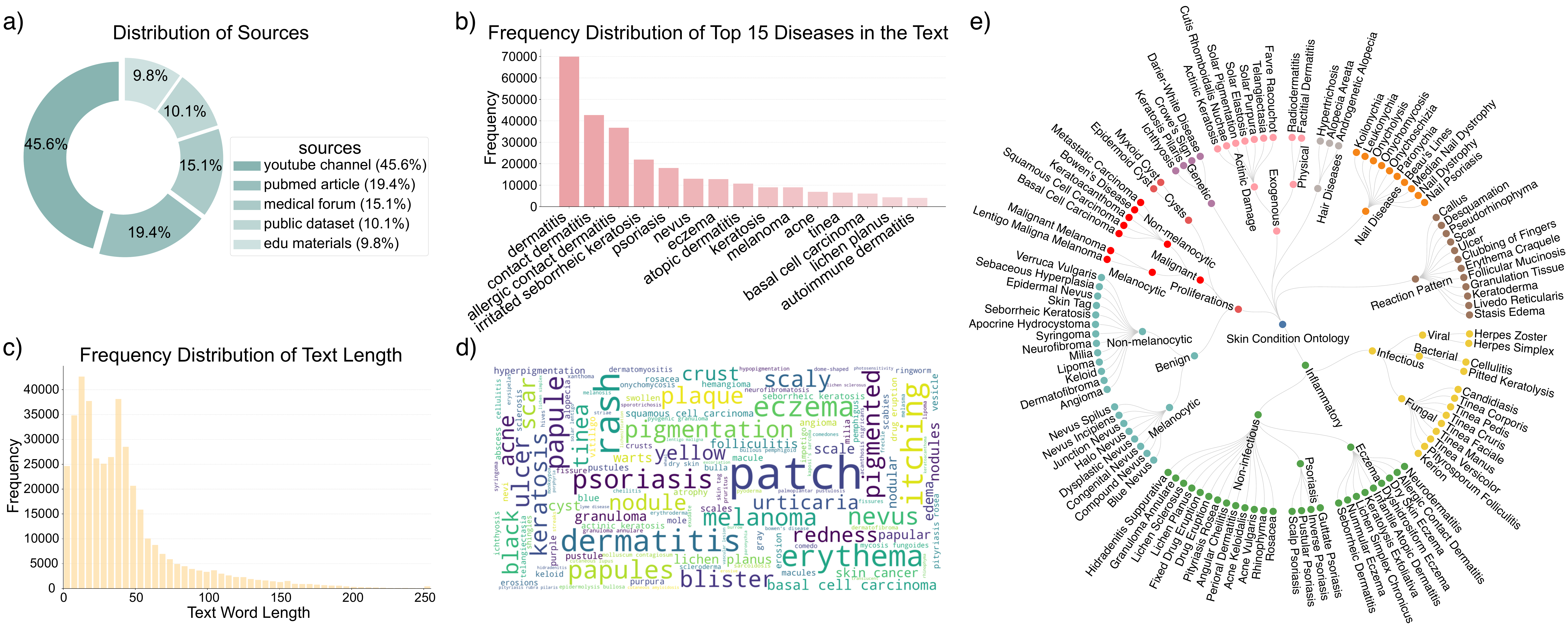}
\vspace{-6mm}
\captionof{figure}{\textbf{Overview of the Derm1M dataset.} The first \textbf{large-scale} vision-language (VL) dataset for dermatology, comprises 1,029,761 skin image-text pairs. a) \textbf{Diverse sources:} Distribution across five types, including YouTube videos, medical forums, PubMed articles, public datasets, and educational materials. b) \textbf{Comprehensive disease coverage:} Frequency of top 15 skin conditions from the total 390 conditions, representing the complex and diverse range of skin diseases encountered in clinical practice. c) \textbf{Rich text descriptions:} Distribution of text lengths across image-text pairs, with a mean length of 41. d) \textbf{Rich contextual information:} Word cloud of common dermatological terms in Derm1M. e) \textbf{Structured domain knowledge:} Expert-developed ontology that conceptually organizes domain knowledge by defining standard entities and their hierarchical disease relationships. Overall, Derm1M advances dermatology AI through unprecedented scale (1M+ image-text pairs, 257× larger than existing VL datasets in dermatology \cite{zhou2024skincap}), comprehensive coverage (390 skin conditions with 130 clinical concepts), and rich clinical information supporting multi-granular learning aligned with clinical practices.}
\label{fig1}
\vspace{-4mm}
\bigskip\bigskip}
\makeatother

\begin{document}

\maketitle
\begin{abstract}
The emergence of vision-language models has transformed medical AI, enabling unprecedented advances in diagnostic capability and clinical applications. However, progress in dermatology has lagged behind other medical domains due to the lack of standard image-text pairs. Existing dermatological datasets are limited in both scale and depth, offering only single-label annotations across a narrow range of diseases instead of rich textual descriptions, and lacking the crucial clinical context needed for real-world applications. To address these limitations, we present Derm1M, the first large-scale vision-language dataset for dermatology, comprising 1,029,761 image-text pairs. Built from diverse educational resources and structured around a standard ontology collaboratively developed by experts, Derm1M provides comprehensive coverage for over 390 skin conditions across four hierarchical levels and 130 clinical concepts with rich contextual information such as medical history, symptoms, and skin tone. To demonstrate Derm1M's potential in advancing both AI research and clinical application, we pretrained a series of CLIP-like models, collectively called DermLIP, on this dataset. The DermLIP family significantly outperforms state-of-the-art foundation models on eight diverse datasets across multiple tasks, including zero-shot skin disease classification, clinical and artifacts concept identification, few-shot/full-shot learning, and cross-modal retrieval. Our dataset and code will be publicly available at https://github.com/SiyuanYan1/Derm1M upon acceptance.

\end{abstract}
  
\vspace{-0.5\baselineskip}
\section{Introduction}
\label{sec:intro}

\begin{table*}[t]
\footnotesize
\centering
\setlength{\tabcolsep}{4pt}  % 设置基础列间距
\begin{tabular}{
    >{\raggedright\arraybackslash}p{0.125\textwidth}|  % 增加第一列宽度
    >{\raggedright\arraybackslash}p{0.13\textwidth}|   % 增加第二列宽度
    >{\centering\arraybackslash}p{0.06\textwidth}
    >{\centering\arraybackslash}p{0.06\textwidth}
    >{\centering\arraybackslash}p{0.07\textwidth}
    >{\centering\arraybackslash}p{0.09\textwidth}
    >{\centering\arraybackslash}p{0.06\textwidth}
    >{\centering\arraybackslash}p{0.07\textwidth}
    >{\centering\arraybackslash}p{0.06\textwidth}
    >{\centering\arraybackslash}p{0.07\textwidth}
}
\hline
\makecell[l]{Dataset\\Type} &
Datasets & 
\makecell{Multi-\\Img.} & 
\makecell{Text\\Desc.} & 
\makecell{Concept\\Label} & 
\makecell{Hierarchical\\Disease} & 
\makecell{Meta\\Info.} & 
\makecell{Dataset\\Size} & 
\makecell{No. of\\Concept} & 
\makecell{No. of\\Condition} \\
\hline
\multirow{4}{*}[-1ex]{\makecell[l]{Classification}}
& HAM10000 \cite{ham10000} & \ding{55} & \ding{55} & \ding{55} & \ding{55} & \ding{51} & 10,015 & 0 & 7 \\
& SD-198 \cite{sd198} & \ding{55} & \ding{55} & \ding{55} & \ding{55} & \ding{55} & 6,584 & 0 & 198 \\
& DermNet & \ding{55} & \ding{55} & \ding{55} & \ding{55} & \ding{55} & 19,500 & 0 & 23 \\
& Derm7pt \cite{derm7pt} & \ding{51} & \ding{55} & \ding{51} & \ding{55} & \ding{55} & 1,711 & 7 & 2 \\
& SkinCon \cite{skincon} & \ding{55} & \ding{55} & \ding{51} & \ding{55} & \ding{55} & 4,346 & 48 & 178 \\
\hline
\multirow{2}{*}{\makecell[l]{Vision-Language}} 
& SkinCAP \cite{zhou2024skincap} & \ding{55} & \ding{51} & \ding{51} & \ding{55} & \ding{55} & 4,000 & 48 & 178 \\
& \cellcolor{gray!20} \textbf{Derm1M (Ours)} & \cellcolor{gray!20}\ding{51} & \cellcolor{gray!20}\ding{51} & \cellcolor{gray!20}\ding{51} & \cellcolor{gray!20}\ding{51} & \cellcolor{gray!20}\ding{51} & \cellcolor{gray!20}\textbf{1,029,761} & \cellcolor{gray!20}\textbf{130} & \cellcolor{gray!20}\textbf{390} \\
\hline
\end{tabular}
\vspace{-2mm}
\caption{\textbf{The comparison among existing public dermatological datasets.} Compared to other datasets, Derm1M provides \textit{multi-modal images} (dermoscopic and clinical images) with \textit{text descriptions}, \textit{concept labels}, and \textit{hierarchical disease labels}, along with comprehensive \textit{meta information}. It contains the largest number of image-text pairs (\textit{Dataset size,} 1,029,761) and covers comprehensive \textit{number of skin conditions} (390) with rich \textit{concept annotations} (130).}
\label{tab1}
\end{table*}

Skin diseases affect 70\% of the global population and rank fourth among nonfatal diseases worldwide \cite{global_burnden}. Despite this prevalence, accurate diagnosis remains challenging, especially where dermatologists are scarce. Primary care physicians, often the first point of contact, demonstrate diagnostic accuracy rates as low as 24\% when confronted with hundreds of possible skin conditions \cite{GP2} ranging from common inflammatory disorders to life-threatening malignancies. This diagnostic challenge stems from dermatology's inherent complexity: physicians must differentiate among numerous conditions, integrate patient information beyond visual examination, and employ standardized clinical concept terminology \cite{dermatology} to communicate their findings and diagnostic reasoning effectively.

Recent advances in computer vision have shown promising results in skin condition diagnosis, achieving superhuman accuracy in controlled settings \cite{dermatologist-level, nmed1, panderm}. However, translating these solutions to clinical practice remains challenging due to limitations in existing datasets. As shown in Table \ref{tab1}, current datasets predominantly focus on narrow condition ranges: HAM10000 \cite{ham10000} contains only 7 classes of pigmented lesions, while even the largest dataset, DermNet \cite{Dermnet}, includes just 19,500 images with 23 condition labels, far below the hundreds of conditions encountered clinically. Additionally, most datasets lack crucial contextual information such as patient history and demographics. Furthermore, reliable dermatology AI requires datasets with dense annotations of physician-understandable concepts and potential dataset bias \cite{skincon,monet,trust}. Clinical concepts like `atypical pigment networks' enhance interpretability, while annotations of dermoscopic artifacts like `dark borders' help identify bias sources. Yet even dedicated concept datasets fall short: SkinCon \cite{skincon}, the largest of its kind, includes just 4,346 images with 48 clinical concepts, failing to capture dermatological diagnosis complexity.

In parallel with these dataset challenges, vision-language models (VLMs) \cite{biomedclip,clip,ophclip,quilt1m} offer a promising direction. These models integrate visual features with clinical descriptions, such as symptom descriptions and clinical findings. Through natural language supervision, they capture contextual information beyond single labels, enabling various applications from diagnosis to visual question answering while providing concept-based explanations \cite{monet} aligned with clinical reasoning. However, their potential for dermatology remains unrealized due to the lack of suitable training data; the only existing dermatological vision-language dataset, SkinCAP \cite{zhou2024skincap}, contains just 4,000 image-text pairs, far below the scale needed for effective pretraining.

To address these limitations, we present Derm1M, the first large-scale vision-language dataset for dermatology with 1,029,761 skin image-text pairs. Recognizing that data quality and diversity are essential, we curated high-quality de-identified dermatological images with detailed textual descriptions from diverse educational sources: 51,309 YouTube videos (10,660 hours), 49,875 medical forum posts, 566,571 PubMed articles, 68 educational materials, and additional two high-quality public datasets. Our comprehensive pipeline (Fig. \ref{fig2}) employs a series of tools such as large language models, speech recognition, object detection, and classifiers to ensure data quality. Unlike noisy web-crawled pairs \cite{biomedclip, monet}, Derm1M is structured around a standard ontology (Fig.~\ref{fig1}e) developed by four senior dermatologists. This expert-guided ontology organizes domain knowledge, enabling hierarchical disease understanding across four hierarchical levels. To the end, our dataset encompasses over 390 skin conditions, 130 clinical concepts, and meta-information on patient history and demographics, representing the richest dermatological knowledge resource to date (Fig.\ref{fig1}, Table \ref{tab1}).

In summary, Derm1M advances dermatology AI by: 
\begin{itemize} 
\item \textbf{Unprecedented scale and clinical relevance:} With over one million image-text pairs (257 times larger than existing vision-language datasets \cite{zhou2024skincap} and 52 times larger than the largest skin condition datasets \cite{Dermnet}), Derm1M organizes comprehensive skin conditions across four hierarchical levels, enabling multi-granular learning aligned with clinical practice.

\item \textbf{Rich contextual information:} Built on comprehensive educational resources and an expert-curated ontology, Derm1M provides 130 clinically relevant concept labels, in contrast to the 48 labels in existing datasets \cite{skincon}. This extensive labeling supports a wide range of applications, from explainable diagnosis to multimodal clinical tasks.

\item \textbf{Benchmark validation:} Our DermLIP model trained on Derm1M outperforms state-of-the-art biomedical foundation models by 9.85\% in zero-shot classification accuracy and 48.1\% in cross-modal retrieval recall, validating Derm1M's effectiveness for multimodal applications.
\end{itemize}

\section{Related Work} \label{sec:related_work} 
\noindent\textbf{Deep Learning in Dermatology.} 
Deep learning in dermatology has evolved from initial transfer learning approaches with ImageNet pretrained CNNs \cite{early, resnet} to public benchmarks like ISIC \cite{isic} and dermatologist-level classification achieved by Esteva et al. \cite{derm} using InceptionV3 \cite{inception}. Recent advancements follow two paradigms: self-supervised learning models (PanDerm \cite{panderm}, SwavDerm \cite{swavderm}) leveraging proprietary million-scale unlabeled images, and vision-language models like MONET \cite{monet} trained on ~100k image-text pairs from PubMed and private textbooks.

\noindent\textbf{Medical Vision-Language Datasets.} Vision-Language models \cite{clip} require extensive high-quality image-text pairs, which remain scarce in dermatology due to the lack of standardized structured reports. Other medical domains have made significant progress: Quilt1M \cite{quilt1m} for pathology using YouTube content demonstrates advancement in that specialty, while broader biomedical datasets like PMC-15M \cite{pmc-clip} and MedTrinity-25M \cite{medtrinity} have collected 10-25 million pairs from web sources like PubMed. However, dermatology has been largely overlooked in these efforts. These broader biomedical datasets contain minimal dermatological content: only ~20k (0.13\%) in PMC-15M and ~10k (0.03\%) in MedTrinity-25M are skin-related pairs, respectively. Furthermore, these data lack the clinical knowledge and structured texts needed to capture dermatology-specific knowledge, making them inadequate for dermatological AI.
\noindent\textbf{Dermatology-specific Vision-Language Datasets.} Prior to our work, SkinCAP \cite{zhou2024skincap} was the only dermatological vision-language dataset, containing merely 4,000 image-text pairs derived from SkinCon \cite{skincon}. This limited scale constrains effective dermatological vision-language model development. Our Derm1M dataset addresses these limitations by integrating an expert-guided ontology, comprehensive disease and clinical context coverage, and diverse educational resources across five data types. By providing 257 times more image-text pairs than SkinCAP and 52 times more than the largest existing skin condition dataset \cite{Dermnet}, Derm1M enables robust vision-language models capable of understanding the complex visual and contextual nuances essential for accurate dermatological assessment.
\section{Derm1M Dataset}
\begin{figure*}[!t]
  \centering
   \includegraphics[width=1\linewidth]{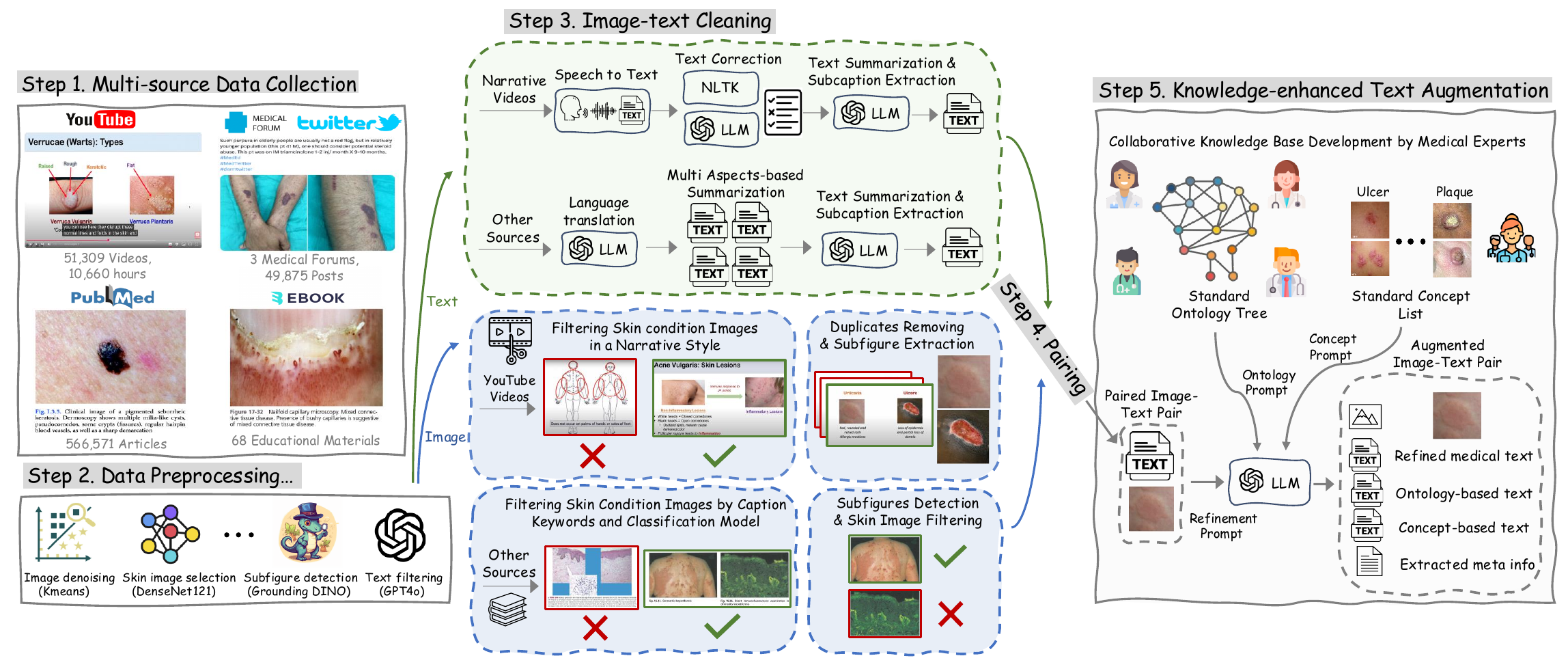}
\vspace{-6mm}
   \caption{\textbf{The overview of the five-stage process for Derm1M dataset construction:} (1) Multi-source data collection, (2) data preprocessing, (3) image-text cleaning, (4) image-text pairing, and (5) knowledge-enhanced text augmentation.}
   \label{fig2}
\end{figure*}

The Derm1M dataset construction process is illustrated in Fig.~\ref{fig2}. Our workflow consists of five steps: (1) collecting dermatological data from diverse educational sources; (2) preprocessing data through various tools; (3) conducting specialized image-text cleaning; (4) establishing accurate image-text pairs through multi-modal alignment; and (5) enriching textual descriptions with domain knowledge using standard ontologies and concepts. This systematic approach ensures both quality and diversity in our dataset.

\subsection{The Construction of Derm1M}
\noindent\textbf{Step 1: Multi-source Data Collection.} To ensure data quality and diversity, we collect primarily from educational resources used by dermatologists and general practitioners (Step 1 of Fig.~\ref{fig2}). We compiled a comprehensive list of dermatology terms from relevant literature and datasets. These terms were systematically used as search queries across multiple data sources, as follows. \noindent \textbf{YouTube.} Our primary data source consists of educational dermatology videos with accompanying narration. We retrieved the top 200 videos for each search query using 355 dermatology terms and manually curated 50 YouTube channels dedicated to dermatology explanations. We prioritized high-resolution videos ($\geq$224p) exceeding 30 seconds, collecting approximately 51k videos. \noindent \textbf{PubMed.} This subset comprises scientific dermatology image-text pairs from medical literature. We retrieved articles (1990-2024) from the PMC Open Access Subset using domain-specific terms, yielding 566,571 articles with approximately 3.6 million images. \noindent \textbf{Medical Forums.} We curated content from dermatology channels across social media platforms, particularly X/Twitter. By reviewing content associated with 58 dermatology-related keywords, we identified 27 relevant channels comprising 14,099 posts. A similar process was employed for other forums including IIYI and Reddit, yielding a total of 49,875 posts. \noindent \textbf{Educational Materials.} We extracted image-text pairs from 68 dermatology educational materials such as clinical guidelines using PyMuPDF, automatically matching images with their nearest caption boxes. When direct extraction was not feasible, Optical Character Recognition (OCR) converted documents into a vectorized format before extraction.
\noindent \textbf{Public Datasets.} We incorporated two public datasets including SCIN \cite{scin} and MSKCC \cite{isic}, creating structured text descriptions by integrating clinical metadata into standardized templates, yielding 17,137 additional image-text pairs.

\noindent\textbf{Step 2: Data Preprocessing.} Step 2 of Fig.~\ref{fig2} shows our preprocessing tools for handling multimodal dermatological data:
(1) \textbf{Image/video processing}: FFmpeg\footnote{https://www.ffmpeg.org/} for keyframe extraction, K-means clustering for denoising, DenseNet121 for skin image selection, DINO \cite{dino} for subfigure detection, EfficientNetV2-S for feature extraction, and PCA for dimension reduction;
(2) \textbf{Audio processing}: Whisper Large-V3 \cite{radford2023robust} for speech-to-text conversion and inaSpeechSegmenter \cite{segmenter} for narration quality assessment;
(3) \textbf{Text processing}: RAKE algorithm for key phrase extraction, OCR for text extraction from images, and NLTK\footnote{https://www.nltk.org/} and GPT-4o for text filtering and refinement.\\

\noindent\textbf{Step 3: Image-Text Cleaning.} As illustrated in Step 3 of Fig.~\ref{fig2}, our cleaning process implements parallel pipelines for image and text content across sources. 

\noindent\textbf{Image Processing.} For YouTube video, we extracted keyframes using FFmpeg by computing inter-frame color histogram differences, with adaptive thresholds based on video length (0.008 for 5-minute videos to 0.25 for 200-minute videos) and applied a DenseNet121 classifier to identify dermatological images. Videos where more than 50\% of keyframes contained dermatological content were retained. We utilized inaSpeechSegmenter \cite{segmenter} to filter out videos lacking sufficient explanatory narration.
For other sources, we implemented hierarchical clustering using EfficientNetV2-S for feature extraction followed by PCA dimension reduction to 50 dimensions. K-means clustering grouped images into 20 major clusters, each further divided into 20 subclusters. Through iterative manual inspections, we refined clusters until only dermatology-related content remained.
For subfigure processing, we trained a DINO \cite{dino} object detector to detect and segment subfigures from composite figures. The detected subfigures were cropped and arranged in a structured left-to-right, top-to-bottom order. A classifier ensured dermatological relevance.

\noindent \textbf{Text Processing.} For YouTube subtitles, we employed Whisper Large-V3 \cite{radford2023robust} for speech-to-text conversion followed by a three-step cleanup: (i) applying RAKE algorithm to identify key phrases and remove stop words; (ii) GPT-4o to correct medical terminology errors; and (iii) generating structured summaries. For \textbf{general text} processing, we implemented language detection and translation for non-English text, filtered out non-medical information blocks, and expanded medical abbreviations contextually. For \textbf{forum} content, we extracted structured multi-aspect summaries of chief complaints and clinical findings; when summarization failed, we extracted symptom and disease entities and connected them with semicolons. Moreover, we enhanced forum text clarity by removing irrelevant elements and eliminating interrogative sentences beginning with phrases like ``What is," while manual inspection removed advertisement-related content. Quality control measures included removing captions with fewer than three words or ten characters and refining OCR-derived text through spell-checking and coherence enhancement.\\

\noindent\textbf{Step 4: Image-Text Pairing.} Our image-text pairing strategies varied by data source:
For \textbf{YouTube}, we defined timestamp intervals between consecutive keyframes as image chunks and transcribed sentences as text chunks. Pairs were formed when temporal overlap exceeded 50\%, with multiple text chunks concatenated when necessary. For unmatched image chunks, we aligned them with the most recent preceding keyframe with associated text.
For \textbf{PubMed} and \textbf{educational materials}, we matched images with captions extracted from XML files or nearest caption boxes. For subfigures, we employed regular expression matching to identify subfigure markers (e.g., A) and (a)), matching each subfigure with its corresponding subcaption.\\

\noindent\textbf{Step 5: Knowledge-Enhanced Text Augmentation.} As shown in Step 5 of Fig.~\ref{fig2}, we developed a systematic approach to enhance raw text with domain knowledge through three sequential processes: standardizing medical terminology, building a comprehensive skin disease ontology, and generating multiple types of knowledge-enhanced text.

\noindent\textbf{Standardizing Terminology.} To enable consistent knowledge representation, we first standardized dermatological terminology by constructing a comprehensive 390-class disease list. This process involved reconciling terminology across multiple datasets and using LLMs to identify and merge disease synonyms. Similarly, we established 130 standardized clinical concepts encompassing morphology, colors, shape characteristics, surface features, distribution patterns, and border properties. These standardized terminologies formed the foundation for our subsequent knowledge augmentation, which aimed to generate three types of enhanced image-text pairs: \textbf{ontology-based text}, \textbf{concept-based text}, and \textbf{refined medical text}. We also extracted patient metadata such as body sites, symptoms, and skin tone to provide contextual enrichment.

\noindent\textbf{Building Skin Disease Ontology.} After standardizing terminology, we developed a hierarchical ontology to formalize relationships between dermatological entities. This knowledge structure was built in two distinct phases. First, four experts collaboratively created a standard ontology (Fig.~\ref{fig1}e) covering 128 skin conditions. Second, we expanded this to incorporate our full set of 390 diseases using an LLM-guided approach (as illustrated in Fig.6, Supplementary). For this expansion phase, we provided the LLM with the standard 128-class ontology tree, a specialized ontology prompt, and our 407-class disease list. Our prompting strategy adhered to four key principles: (1) preservation of the original hierarchy; (2) accurate placement of each disease; (3) reasoned justification for each placement; and (4) systematic handling of diseases with uncertain classification. Through this approach, the LLM integrated additional diseases while maintaining hierarchical integrity. We conducted five iterations with manual validation between each round, ultimately producing a comprehensive ontology tree containing 371 skin diseases, while 19 diseases remained unplaced due to classification uncertainty.

\noindent\textbf{Generating Enhanced Text.} With our comprehensive ontology established, we proceeded to generate three types of knowledge-enhanced textual descriptions. First, we generate \textbf{ontology-based text} by identifying diseases finally diagnosed in the original clinical descriptions, retrieving their hierarchical paths from our ontology (e.g., folliculitis: inflammatory → infectious → bacterial → folliculitis), and transforming these relationships into structured textual descriptions.  Next, for the \textbf{concept-based text}, we utilized our 130-item concept list to extract visual characteristics from the raw medical text. Finally, we generated the \textbf{refined medical text} by integrating both the ontology-based and concept-based texts into the original descriptions, creating comprehensive knowledge-enhanced representations that leverage hierarchical disease relationships and standardized visual concepts with context information such as medical history, symptoms, body sites, and treatment. 

Please refer to Section 1, Fig. 1-6 in the Supplementary for more detailed data curation pipelines of Derm1M.

\begin{table*}[t] 
\footnotesize 
\centering 
\setlength{\tabcolsep}{2.2pt} 
\begin{tabular}{c|l|cc|cccc|c} 
\hline 
\multirow{3}{*}{\begin{tabular}[c]{@{}c@{}}\textbf{Type}\end{tabular}} & \multirow{3}{*}{\textbf{Method}} & \multicolumn{2}{c|}{\textbf{Encoder}} & \multicolumn{5}{c}{\textbf{Accuracy}} \\ 
\cline{3-9} 
& & \multirow{2}{*}{\textbf{Vision}} & \multirow{2}{*}{\textbf{Text}} & \textbf{HAM \cite{ham10000}} & \textbf{F17K \cite{f17k}} & \textbf{PAD \cite{pad}} & \textbf{Daffodil \cite{Daffodil}} & \multirow{2}{*}{\textbf{Average}} \\ 
& & & & \textbf{(7 classes)} & \textbf{(113 classes)} & \textbf{(6 classes)} & \textbf{(5 classes)} & \\ 
\hline 
\multirow{3}{*}{\begin{tabular}[c]{@{}c@{}}General\\VLMs\end{tabular}} & CLIP-B16 \cite{clip} & ViT-B16 & GPT77 & .2023 & .0653 & .4555 & .4534 & .2941 \\ 
& SigLIP \cite{siglip} & ViT-B16 & SigLIP64 & .2502 & .1007 & .5315 & .6995 & .3955 \\ 
& CoCa \cite{yu2022coca} & ViT-B32 & GPT77 & .1796 & .0647 & .3623 & .5424 & .2873 \\ 
\hline 
\multirow{3}{*}{\begin{tabular}[c]{@{}c@{}}Biomedical\\VLMs\end{tabular}} & PMC-CLIP \cite{pmc-clip} & ResNet50 & PMB256 & .5349 & .0302 & .4273 & .3529 & .3363 \\ 
& BiomedCLIP \cite{biomedclip} & ViT-B16 & PMB256 & .6347 & .0955 & .4512 & .5817 & .4408 \\ 
& MONET \cite{monet} & ViT-L14 & GPT77 & .2967 & .1397 & .4425 & .7215 & .4001 \\ 
\hline 
\multirow{4}{*}{\begin{tabular}[c]{@{}c@{}}DermLIP\\Variants\end{tabular}} & \cellcolor{gray!15}DermLIP & \cellcolor{gray!15}ViT-B16 & \cellcolor{gray!15}GPT77 & \cellcolor{gray!15}\textbf{.6820} & \cellcolor{gray!15}.2278 & \cellcolor{gray!15}.6074 & \cellcolor{gray!15}.7257 & \cellcolor{gray!15}.5607 \\ 
& \cellcolor{gray!15}DermLIP & \cellcolor{gray!15}ViT-B16 & \cellcolor{gray!15}BMB77 & \cellcolor{gray!15}.6647 & \cellcolor{gray!15}.2740 & \cellcolor{gray!15}.6095 & \cellcolor{gray!15}.7497 & \cellcolor{gray!15}.5745 \\ 
& \cellcolor{gray!15}DermLIP & \cellcolor{gray!15}ViT-B16 & \cellcolor{gray!15}PMB256 & \cellcolor{gray!15}.5236 & \cellcolor{gray!15}.3133 & \cellcolor{gray!15}.6161 & \cellcolor{gray!15}.7702 & \cellcolor{gray!15}.5558 \\ 
& \cellcolor{gray!15}DermLIP & \cellcolor{gray!15}PanDerm-B & \cellcolor{gray!15}PMB256 & \cellcolor{gray!15}.6274 & \cellcolor{gray!15}\textbf{.3162} & \cellcolor{gray!15}\textbf{.6247} & \cellcolor{gray!15}\textbf{.7832} & \cellcolor{gray!15}\textbf{.5879} \\ 
\hline 
\end{tabular} 
\vspace{-2mm} 
\caption{Benchmarking results on Zero-shot image classification (Acc).} 
\label{tab:zs} 
\end{table*}

\begin{table*}[t]
\footnotesize
\centering
\setlength{\tabcolsep}{2.5pt}  % 缩小列间距以适应额外的列
\begin{tabular}{l|cc|ccc|ccc|ccc|ccc|ccc}
\hline
\multirow{2}{*}{\textbf{Method}} & \multicolumn{2}{c|}{\textbf{Encoder}} & \multicolumn{3}{c|}{\textbf{HAM (7 classes)}} & \multicolumn{3}{c|}{\textbf{F17K (113 classes)}} & \multicolumn{3}{c|}{\textbf{PAD (6 classes)}} & \multicolumn{3}{c|}{\textbf{Daffodil (5 classes)}} & \multicolumn{3}{c}{\textbf{Average}} \\
\cline{2-18}
 & \textbf{Vision} & \textbf{Text} & \textbf{1\%} & \textbf{10\%} & \textbf{100\%} & \textbf{1\%} & \textbf{10\%} & \textbf{100\%} & \textbf{1\%} & \textbf{10\%} & \textbf{100\%} & \textbf{1\%} & \textbf{10\%} & \textbf{100\%} & \textbf{1\%} & \textbf{10\%} & \textbf{100\%} \\
\hline
CLIP-B16 \cite{clip} & ViT-B16 & GPT77 & .697 & .771 & .823 & .101 & .242 & .407 & .482 & .607 & .718 & .767 & .911 & .942 & .512 & .633 & .723 \\
SigLIP \cite{siglip} & ViT-B16 & SigLIP64 & .638 & .731 & .800 & .058 & .161 & .266 & .430 & .538 & .586 & .572 & .775 & .879 & .425 & .551 & .633 \\
CoCa \cite{yu2022coca} & ViT-B32 & GPT77 & .684 & .698 & .687 & .090 & .219 & .402 & .516 & .607 & .633 & .715 & .801 & .806 & .501 & .581 & .632 \\
\hline
PMC-CLIP \cite{pmc-clip} & ResNet50 & GPT77 & .707 & .781 & \underline{.855} & .114 & .275 & .451 & \underline{.534} & .594 & .725 & .744 & .915 & .964 & .525 & .641 & .749 \\
BiomedCLIP \cite{biomedclip} & ViT-B16 & PMB256 & .705 & .697 & .687 & .100 & .182 & .335 & .501 & .640 & .679 & .725 & .811 & .819 & .508 & .583 & .630 \\
MONET \cite{monet} & ViT-L14 & GPT77 & .707 & .800 & .850 & .120 & .294 & .471 & .505 & .651 & .727 & .789 & .931 & .963 & .530 & .669 & .753 \\
BiomedGPT-B \cite{biomedgpt} & GPT & GPT & \textbf{.774} & .803 & .847 & .036 & .118 & .242 & .395 & .445 & .503 & .478 & .691 & .793 & .421 & .514 & .596 \\
PanDerm-B \cite{panderm} & ViT-B16 & - & .703 & .774 & .854 & .093 & .210 & .463 & .471 & .610 & .694 & .736 & .918 & \underline{.978} & .501 & .628 & .747 \\
\color{gray!80}PanDerm-L \cite{panderm} & \color{gray!80}ViT-L14 & \color{gray!80}- & \color{gray!80}.711 & \color{gray!80}.790 & \color{gray!80}.878 & \color{gray!80}.138 & \color{gray!80}.343 & \color{gray!80}.584 & \color{gray!80}.587 & \color{gray!80}.651 & \color{gray!80}.759 & \color{gray!80}.833 & \color{gray!80}.955 & \color{gray!80}.987 & \color{gray!80}.567 & \color{gray!80}.685 & \color{gray!80}.802 \\
\hline
\rowcolor{gray!15}
DermLIP & ViT-B16 & GPT77 & .746 &\underline{ .811} & .852 & \textbf{.160} & \underline{.356} & \underline{.510} & .518 & \textbf{.659} & \textbf{.761} & \textbf{.855} & \underline{.937} & .964 & \underline{.570} & \underline{.691} & \underline{.772} \\
\rowcolor{gray!15}
DermLIP & PanDerm-B & PMB256 & \underline{.755} & \textbf{.817} & \textbf{.881} & \underline{.158} & \textbf{.367} & \textbf{.570} & \textbf{.566} & \underline{.653} & \underline{.740} & \underline{.846} & \textbf{.948} & \textbf{.982} & \textbf{.586} & \textbf{.696} & \textbf{.793} \\
\hline
\end{tabular}
\vspace{-2mm}
\caption{Benchmarking results on Linear Evaluation (Acc) across different label ratios. The best-performing model for each setting is in bold, and the second-best is underlined.}
\label{tab:lp}
\end{table*}

\subsection{Dataset Statistics and Analysis}
The resulting Derm1M dataset contains 1,029,761 high-quality dermatological image-text pairs, comprising 403,563 images with refined medical text, 403,563 ontology-based text, and 222,635 concept-based text. Spanning 390 skin conditions and 130 standardized visual concepts, our dataset sources include YouTube channels (45.6\%), PubMed articles (19.4\%), medical forums (15.1\%), public datasets (10.1\%), and educational materials (9.8\%) as shown in Fig.~\ref{fig1}a. Fig.~\ref{fig1}b presents the top 15 diseases in Derm1M, with a distribution reflecting clinical practice where inflammatory dermatoses predominate while skin cancers remain diagnostically significant. The dataset features rich descriptions averaging 41 tokens per caption (Fig.~\ref{fig1}c), with disease terms and clinical concepts comprising the main vocabulary (Fig.~\ref{fig1}d). A distinguishing feature of Derm1M is its ontological structure (Fig.~\ref{fig1}e), which hierarchically organizes 371 skin conditions (91.1\% of all diseases in Derm1M) across inflammatory conditions, reaction patterns, tumors, and other categories. This comprehensive organization facilitates fine-grained disease relationship modeling to support clinical diagnostic AI systems. Please refer to Section 2 in the Supplementary for additional dataset statistics (Fig. 7-9), visual examples of image-text pairs (Fig. 10-13), and the complete lists of 130 concepts (Table 5) and 390 skin conditions (Table 6).

\section{Experiments with Derm1M}
\subsection{Experimental Setup}
\noindent\textbf{Downstream Datasets.}
For downstream evaluation, we follow CLIP \cite{clip} and evaluate our model across four task categories using eight datasets. (1) \textbf{Zero-shot disease classification}, comprising dermoscopic and clinical image-based seven and six class skin cancer diagnosis (HAM10000 \cite{ham10000}, PAD-UFES-20 \cite{pad}), 113 class general skin condition classification (Fitzpatrick17K \cite{f17k}), and 5 class non-melanoma skin disease classification including rare diseases (Daffodil \cite{Daffodil}). (2) \textbf{Few-shot, full-shot learning under linear evaluation}, using the same four datasets with 1\%, 10\%, and 100\% labeled training data. (3) \textbf{Zero-shot concept identification} for identifying concepts, including general skin condition-related 32 clinical concepts (SkinCon \cite{skincon}), melanoma-related 7-point checklist clinical concepts (Derm7pt \cite{derm7pt}), and artifacts-related concepts (ISIC \cite{isic}). (4) \textbf{Cross-modal retrieval} for image-to-text and text-to-image retrieval using Derm1M holdout test set and SkinCAP \cite{zhou2024skincap}. Please refer to Section 5, Table 7 and 8 in the Supplementary for more details.\\

\noindent\textbf{Baselines.}
We compare three types of models. For zero-shot learning including skin disease classification, concept identification, and cross-modal retrieval, we include (1) \textbf{General VLMs} such as CLIP \cite{clip}, SigLIP \cite{siglip}, and CoCa \cite{yu2022coca}. (2) \textbf{Biomedical VLMs}, including PMC-CLIP \cite{pmc-clip} and BiomedCLIP \cite{biomedclip}, which are two SOTA biomedical foundation models. We also compare with MONET \cite{monet}, a dermatology VLM designed for concept identification. (3) \textbf{DermLIP variants}, which include our DermLIP trained using CLIP's contrastive learning objective on Derm1M with different vision and text encoders. For vision encoders, we choose ViT-B16 and PanDerm-B. The latter is the base version of a self-supervised dermatology foundation model \cite{panderm}. For text encoders, we evaluate GPT2 \cite{gpt2} with a context length of 77 (GPT77), BiomedBert (BMB77) \cite{biomedbert} with a context length of 77, and PubmedBert (PMB256) \cite{pubmedbert} with a context length of 256. For few-shot and full-shot learning, we additionally compare with two biomedical foundation models, BiomedGPT-B \cite{biomedgpt}, a generalist biomedical foundation model, and PanDerm-B \cite{panderm}, a self-supervised dermatology foundation model. To ensure fair comparison across all baselines, we use the ViT-base or encoders with similar parameter sizes when available, with exceptions for PMC-CLIP using ResNet50 \cite{resnet} and MONET using ViT-large, as these are their only configurations.

\begin{figure*}[t]
    \centering
    \setlength{\tabcolsep}{5pt}  % Adjust this value to get desired spacing
    \renewcommand{\arraystretch}{0.8}
    \begin{tabular}{c@{\hspace{10pt}}c}
        \includegraphics[width=0.48\linewidth]{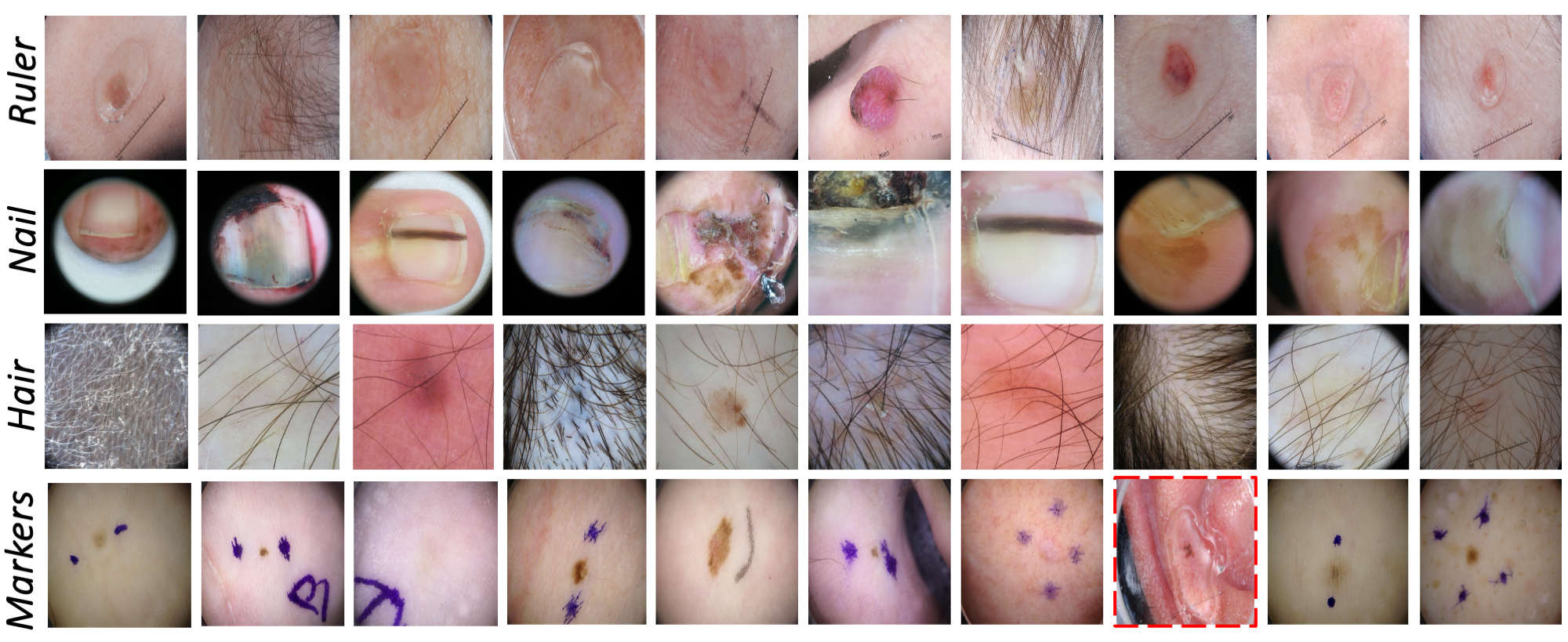} &
        \includegraphics[width=0.48\linewidth]{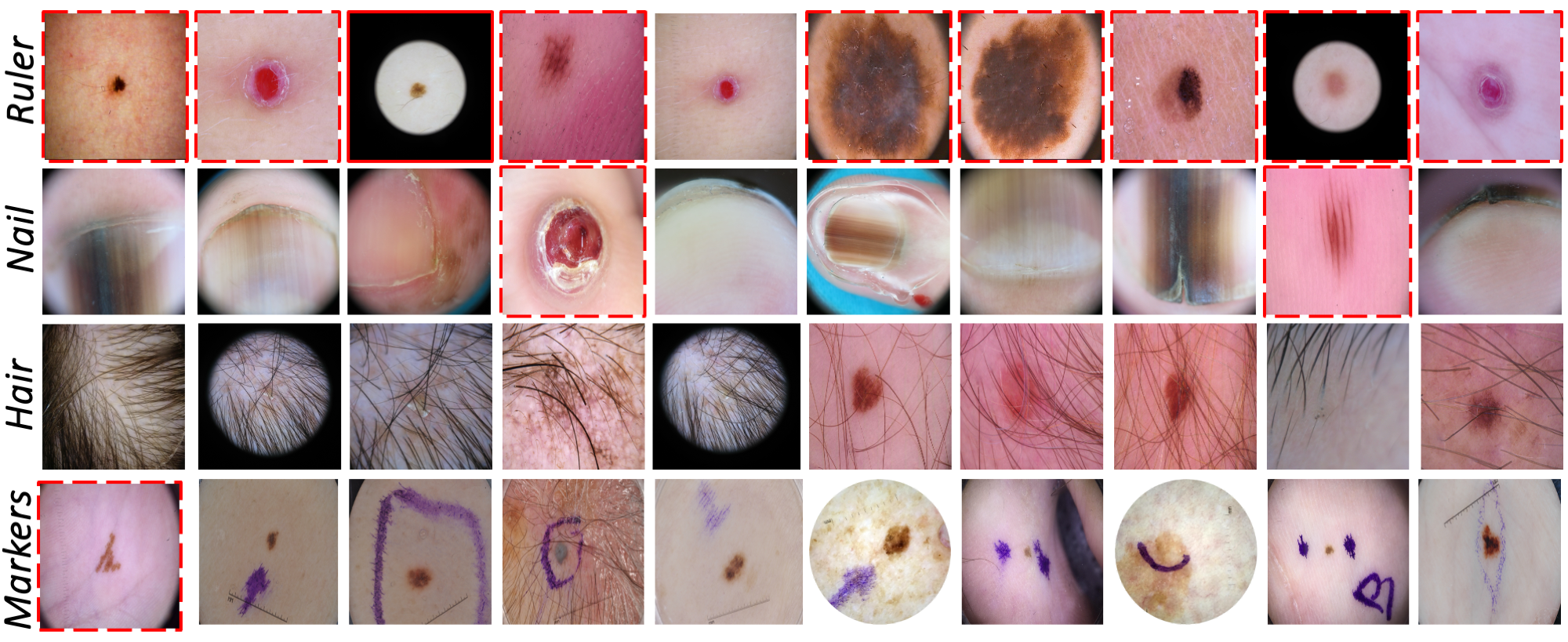} \\
        \footnotesize{\small(a) DermLIP (Ours)} & \footnotesize{\small(b) MONET \cite{monet}.} \\
    \end{tabular}
\vspace{-3mm}
\caption{\small \textbf{Comparison of DermLIP and MONET \cite{monet} on artifact concept detection in the ISIC dataset \cite{isic}.} The top 10 images with the highest scores for common dermatological artifacts (ruler, nail, hair, markers). Red boxes highlight incorrect concept annotations.}
\label{fig4}
\end{figure*}

\subsection{Benchmark Results}
This section evaluates the effectiveness of Derm1M dataset by training a series of CLIP-like models with varied encoders. It compares our DermLIP model against existing general and biomedical VLMs, providing insights for future dermatology-specific VLM improvements. We also include ablation studies about different training strategies in Section 4 and Table 1-3 in the Supplementary.

\noindent\textbf{Zero-shot Skin Disease Classification.} 
Clinical dermatology spans hundreds of skin conditions, making zero-shot classification valuable for identifying unseen diseases without additional training. We evaluated DermLIP's zero-shot performance across four external datasets by converting class labels into sentence templates (e.g., ``melanoma" into ``This is a dermatological image of melanoma"). Details on prompting strategies are in Section 5 of the Supplementary.

As shown in Table~\ref{tab:zs}, DermLIP variants substantially outperform both general VLMs and biomedical VLMs across all test sets. DermLIP with ViT-B16/GPT77 exceeds the next-best baseline (BiomedCLIP) by significant margins: 13.23\% on the Fitzpatrick17K dataset (F17K), 15.62\% on the PAD dataset, and 11.99\% higher average accuracy across all four datasets. Further analysis of encoder combinations revealed that DermLIP with PanDerm-B/PMB256 achieves the best performance. This configuration improves upon DermLIP with ViT-B16/GPT77 by 2.72\% in average accuracy, with notable gains on F17K (8.84\%) and Daffodil (5.8\%). These results highlight two critical factors for enhancing zero-shot capability: (1) the value of domain-specific pre-trained encoders for both vision and text modalities, and (2) the importance of supporting longer text context lengths, as PubMedBert's 256-token capacity better accommodates the extended textual descriptions present in our dataset (as illustrated in Fig.~\ref{fig1}c), compared to the 77-token limitation of BiomedBert and GPT77.

 \noindent\textbf{Few-shot and Full-shot Skin Disease Classification with Linear Evaluation.} We evaluated DermLIP through linear probing across varying data availability scenarios: few-shot (1\% and 10\% of training data) and full-shot (100\%). As shown in Table~\ref{tab:lp}, DermLIP variants consistently outperform existing models, with DermLIP (PanDerm-B/PMB256) achieving the best overall results across all scenarios. In the data-scarce scenario (1\% of training data), DermLIP with PanDerm-B/PMB256 achieves 58.6\% average accuracy, outperforming the next-best baseline MONET by 5.6\%. This advantage persists as training data increases, with DermLIP (PanDerm-B/PMB256) outperforming MONET by 2.7\% and 4.0\% in the 10\% and 100\% scenarios. Notably, while MONET uses a larger vision encoder (ViT-L14 vs. ViT-B16), our dataset enables DermLIP to achieve better results with smaller models. When comparing encoder configurations, DermLIP (PanDerm-B/PMB256) improves upon DermLIP (ViT-B16/GPT77) by 1.6\%, 0.5\%, and 2.1\% in average accuracy across all four datasets at 1\%, 10\%, and 100\% data scenarios, respectively. These findings align with our zero-shot results, demonstrating the importance of domain-specific pre-training for both visual and textual encoders.

\begin{figure}[!t]
  \begin{minipage}{0.65\linewidth}
    \centering
    \includegraphics[width=\linewidth]{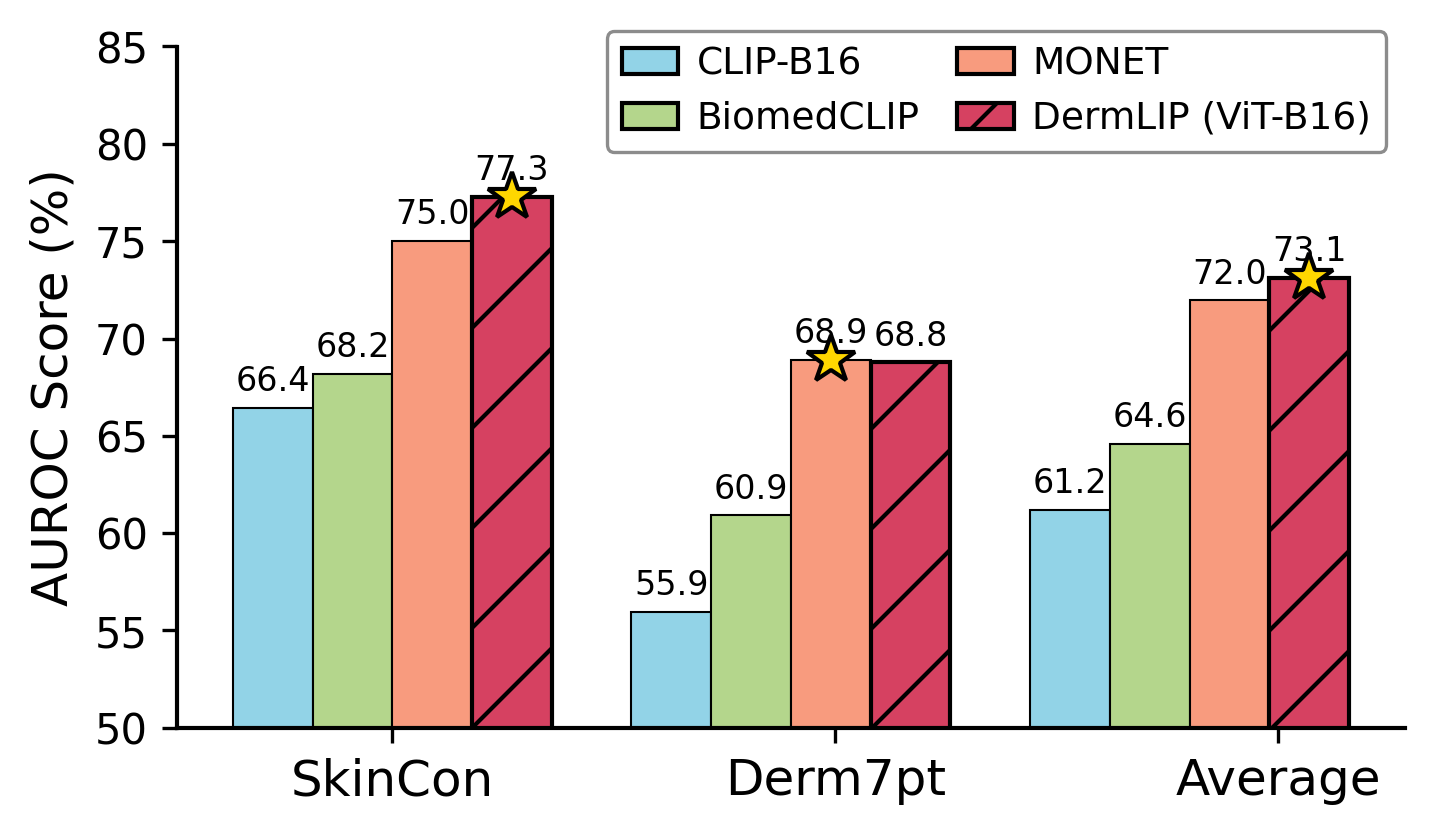}
  \end{minipage}%
  \begin{minipage}{0.3\linewidth}
    \caption{Benchmarking on zero-shot clinical concept identification for SkinCon and Derm7pt datasets.}
    \label{fig3}
  \end{minipage}
\end{figure}
\begin{table*}[t]
\footnotesize
\centering
\setlength{\tabcolsep}{4.5pt}
\begin{tabular}{c|l|cc|cccc|cccc}
\hline
\multirow{3}{*}{\textbf{Type}} & \multirow{3}{*}{\textbf{Method}} & \multicolumn{2}{c|}{\textbf{Encoder}} & \multicolumn{8}{c}{\textbf{Cross-modal Retrieval}} \\
\cline{3-12}
 & & \multirow{2}{*}{\textbf{Vision}} & \multirow{2}{*}{\textbf{Text}} & \multicolumn{4}{c|}{\textbf{Holdout (n=9806)}} & \multicolumn{4}{c}{\textbf{SkinCAP (n=3989)}} \\
\cline{5-12}
 & & & & \multicolumn{2}{c}{\textbf{I2T}} & \multicolumn{2}{c|}{\textbf{T2I}} & \multicolumn{2}{c}{\textbf{I2T}} & \multicolumn{2}{c}{\textbf{T2I}} \\
 & & & & \textbf{R@10} & \textbf{R@50} & \textbf{R@10} & \textbf{R@50} & \textbf{R@10} & \textbf{R@50} & \textbf{R@10} & \textbf{R@50} \\
\hline
\multirow{3}{*}{\centering General VLMs} 
 & CLIP-B16 \cite{clip} & ViT-B16 & GPT77 & .0713 & .1542 & .0594 & .1386 & .0913 & .2354 & .0592 & .1860 \\
 & SigLIP \cite{siglip} & ViT-B16 & SigLIP64 & .1184 & .2314 & .1109 & .2221 & .1329 & .3184 & .0955 & .2585 \\
 & CoCa \cite{yu2022coca} & ViT-B32 & GPT77 & .0669 & .1477 & .0514 & .1244 & .0857 & .2351 & .0682 & .1985 \\
\hline
\multirow{3}{*}{\centering Biomedical VLMs} 
 & PMC-CLIP \cite{pmc-clip} & ViT-L14 & GPT77 & .0751 & .1589 & .0627 & .1381 & .0672 & .1908 & .0649 & .1855 \\
 & BiomedCLIP \cite{biomedclip} & ViT-B16 & PMB256 & .1657 & .2789 & .1541 & .2761 & .1359 & .3429 & .1238 & .3304 \\
 & MONET \cite{monet} & ViT-L14 & GPT77 & .1290 & .2509 & .1215 & .2569 & .1421 & .3384 & .1492 & .3490 \\
\hline
\multirow{2}{*}{\centering DermLIP Variants} &
\cellcolor{gray!15}DermLIP & \cellcolor{gray!15}ViT-B16 & \cellcolor{gray!15}GPT77 & \cellcolor{gray!15}.4069 & \cellcolor{gray!15}.6021 & \cellcolor{gray!15}.3966 & \cellcolor{gray!15}.5992 & \cellcolor{gray!15}.1567 & \cellcolor{gray!15}.3632 & \cellcolor{gray!15}.1594 & \cellcolor{gray!15}.3567 \\
 & \cellcolor{gray!15}DermLIP & \cellcolor{gray!15}PanDerm-B & \cellcolor{gray!15}PMB256 & \cellcolor{gray!15}\textbf{.5998} & \cellcolor{gray!15}\textbf{.7610} & \cellcolor{gray!15}\textbf{.5930} & \cellcolor{gray!15}\textbf{.7605} & \cellcolor{gray!15}\textbf{.2018} & \cellcolor{gray!15}\textbf{.4327} & \cellcolor{gray!15}\textbf{.2128} & \cellcolor{gray!15}\textbf{.4620} \\
\hline
\end{tabular}
\vspace{-2mm}
\caption{Cross-modal retrieval results. I2T represents image-to-text retrieval and T2I represents text-to-image retrieval.}
\label{tab:retrieval}
\end{table*}
\noindent\textbf{Zero-shot Concept Identification.} Beyond disease classification, we evaluate DermLIP's capability to identify concepts in dermatological images. We first compare it with MONET \cite{monet}, a concept-specialized VLM, for detecting non-clinical artifacts that may introduce bias in model predictions. Following MONET's approach on the ISIC dataset \cite{isic}, Fig.~\ref{fig4} shows the top 10 dermoscopic images with the highest concept scores for various dermoscopic artifacts as identified by DermLIP and MONET. As seen in the figure, DermLIP accurately identifies rulers, nails, hair, and markers, whereas MONET struggles with subtle rulers and performs worse on nail images. We further evaluated DermLIP's zero-shot clinical concept identification on SkinCon \cite{skincon} (32 concepts) and Derm7pt \cite{derm7pt} (7 classes). As shown in Fig.~\ref{fig3}, DermLIP (ViT-B16/GPT77) achieves an AUROC of 73.1\%, outperforming MONET by 1.1\%. On the SkinCon dataset, DermLIP improves AUROC by 2.3\%, while it performs comparably on Derm7pt (68.8\% vs. 68.9\%). These results demonstrate DermLIP's versatility across dermatological tasks, as it outperforms even larger specialized VLMs in dermatology. By accurately identifying both clinical concepts and artifacts, DermLIP can facilitate dataset error auditing and improve model explainability.

\noindent\textbf{Cross-modal Retrieval.} Cross-modal retrieval enables searching for relevant images using text descriptions and vice versa, which is valuable for dermatology education and clinical decision support. We evaluated DermLIP on two datasets: the internal Derm1M holdout set and the external SkinCAP dataset \cite{zhou2024skincap}. As shown in Table~\ref{tab:retrieval}, DermLIP variants significantly outperform all baselines. On the holdout set, DermLIP (PanDerm-B/PMB256) achieves Recall@10 scores of 59.98\% (I2T) and 59.30\% (T2I), outperforming the next-best baseline BiomedCLIP by over 43\%. Similar advantages persist on the external SkinCAP dataset, with DermLIP (PanDerm-B/PMB256) outperforming MONET by 5.97\% (I2T) and 6.36\% (T2I). DermLIP (PanDerm-B/PMB256) also significantly outperforms DermLIP (ViT-B16/GPT77) across all metrics, confirming the advantage of domain-specific encoders for retrieval tasks, aligning with our findings in disease classification tasks.

% {\small\begin{verbatim}
%    \usepackage{graphicx} ...
%    \includegraphics[width=0.8\linewidth]
%                    {myfile.pdf}
% \end{verbatim}
% }

\section{Potential Applications of Derm1M}
The unique features of Derm1M unlock many potential applications. We introduce several of them and leave implementation for future work.

\noindent\textbf{Model explanation}. Concept Bottleneck Models (CBMs) \cite{cbm} are interpretable AI approaches that predict class labels by first projecting images into human-understandable concepts. This requires datasets with both class labels and concept annotations, which are expensive to create in medicine. Label-free CBMs \cite{label-free,labo} leverage CLIP's cross-modal alignment to generate concepts without manual annotations, reducing the labeling burden. However, CLIP struggles with dermatology applications due to domain gaps. Our dataset's 130 clinical concepts enable training dermatology-specific CLIP that generates more precise clinical concepts for CBM development. Unlike previous datasets with limited concept coverage (Table~\ref{tab1}), Derm1M provides comprehensive concept coverage aligned with clinical diagnostic workflows, enhancing model interpretability.

\noindent\textbf{Fine-grained error analysis}. Identifying data and model mistakes is crucial for deploying medical AI systems \cite{monet}. Derm1M's concept coverage enables systematic data error auditing \cite{domino,monet} to identify performance disparities across demographic groups and potential biases. Model auditing \cite{monet} can detect reasoning errors and knowledge gaps in deployed systems. By utilizing frameworks like \cite{monet, describe,trust,domino}, researchers can identify when models make predictions for incorrect reasons, ensuring equitable performance across diverse populations. This capability is crucial for transparent and responsible clinical implementation.

\noindent\textbf{Building foundation models}. The unprecedented scale of Derm1M enables the training of specialized vision-language foundation models using larger parameter architectures for dermatology. These models can serve as generalist models for various multimodal dermatological tasks, similar to those developed in other medical domains \cite{echoclip,conch}. Additionally, our dataset can be structured into instruction-following formats to fine-tune multimodal large language models (e.g., LLLaVA Med \cite{llava_med}) facilitating the creation of dermatology-specific AI assistants that enhance clinical decision-making.

\noindent\textbf{Multimodal tasks}. Derm1M supports visual question answering similar to BiomedCLIP \cite{biomedclip} methodology. Models can leverage the dataset's rich context to respond to complex queries about skin images, such as lesion characteristics, skin tone analysis, and differential diagnoses that require integrating visual features with clinical knowledge.

\noindent\textbf{Clinical applications}. Derm1M enables differential diagnosis of 390 skin conditions, significantly exceeding existing datasets \cite{f17k,sd198,Dermnet} in both class coverage and data size. It is also the first public dataset to support hierarchical skin condition classification with a taxonomic structure that consists of four levels, from broad categories to specific subtypes. This structure mirrors clinical diagnostic workflows and enables AI systems to deliver assessments at varying levels of specificity.

\section{Conclusion and Limitations} In this work, we introduce Derm1M, the first large-scale vision-language dataset for dermatology. Our DermLIP models trained on this dataset significantly outperform existing foundation models across diverse tasks, demonstrating Derm1M's potential. A limitation of Derm1M is the noise introduced from educational resources—a challenge widely shared by representative biomedical vision-language datasets \cite{biomedclip,pmc-clip,quilt1m,monet}. We alleviated this through rigorous data processing and ontology knowledge augmentation. Despite these challenges, Derm1M stands as a pioneering contribution to the field, addressing a critical gap where no comparable dataset existed and will accelerate advances in multimodal AI applications for dermatology.

{
    \small
    \bibliographystyle{ieeenat_fullname}
    \bibliography{main}
}
 
% % WARNING: do not forget to delete the supplementary pages from your submission 
% \clearpage
% \setcounter{page}{1}
% \maketitlesupplementary

% \maketitle
% \begin{abstract}
% In this supplementary material, we present the following contents: (1) more detailed data curation pipelines. (2) additional dataset statistics. (3) downstream dataset details. (4) additional ablation studies, and (5) additional implementation details. 

% \end{abstract}

\section{More Detailed Data Curation Pipelines}

\subsection{YouTube Video}
\begin{figure}[t]
  \centering
  \includegraphics[width=\linewidth]{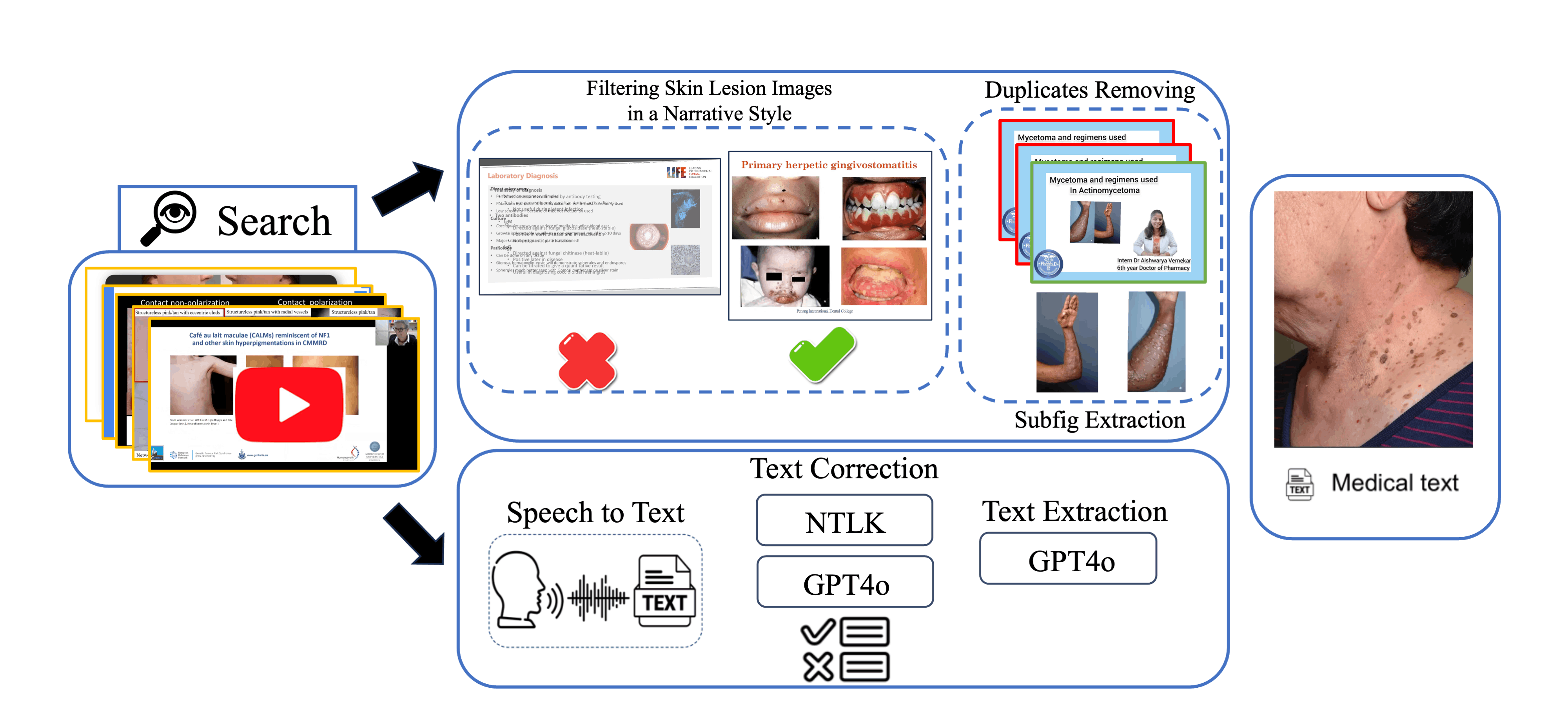}
  \caption{\textbf{Curation pipeline for YouTube content.} Our process begins with searching and collecting 51k videos from educational channels, followed by filtering to identify narrative-style content with high-quality explanations. We then extract and denoise text using a combination of speech-to-text models, handcrafted algorithms, and LLMs. Finally, we align the processed text with corresponding image pairs to create a curated dataset.}
  \label{fig_youtube1}
\end{figure}

\begin{figure*}[t]
  \centering
  \includegraphics[width=0.9\linewidth]{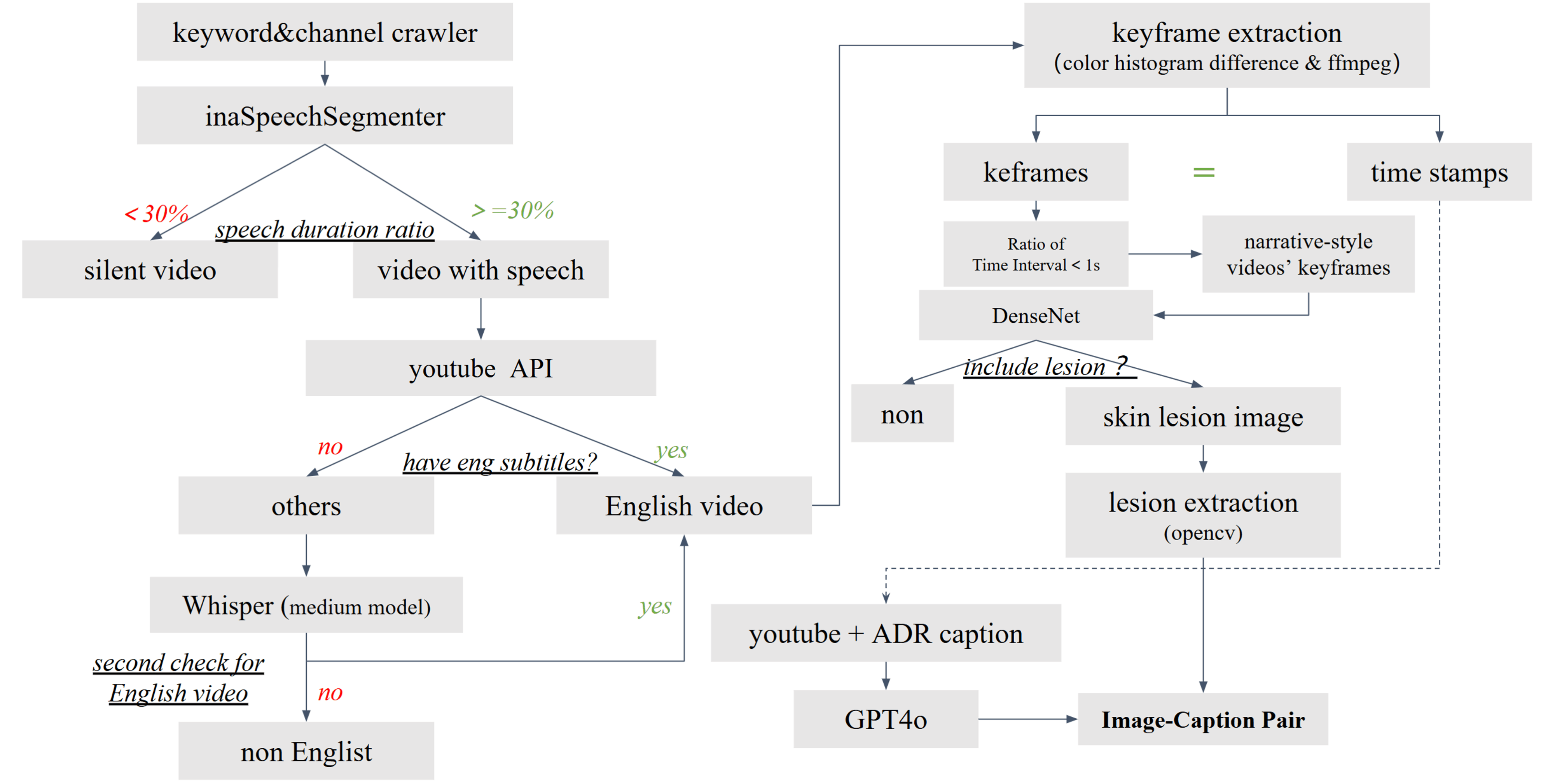}
  \caption{Flow chart of the curation pipeline for YouTube content.}
  \label{fig_youtube2}
\end{figure*}

The main curation pipeline for YouTube is illustrated in Fig.~\ref{fig_youtube1} and Fig.~\ref{fig_youtube2} and detailed below.

\noindent \textbf{Collecting Representative Channels and Videos.}
We constructed a comprehensive list of over 355 dermatology-related terms by consulting relevant literature and publicly available datasets. These terms encompass skin disease-related concepts, common names for various skin conditions, and associated synonyms. For each keyword, we retrieved the top 200 recommended videos from search queries. Additionally, based on our empirical observation that channel-based searches yield more focused and higher-quality explanatory videos compared to keyword searches, we manually curated 50 YouTube channels dedicated to dermatology explanations and downloaded their videos. During the downloading process, we prioritized the highest-resolution version of each video while filtering out videos shorter than 30 seconds or with a resolution below 224p. In total, we collected approximately 51k videos.

\noindent \textbf{Filtering for Narrative-Style Videos.}
We assessed each video to determine: (1) whether it contains a sufficient number of usable dermatological images, and (2) whether it qualifies as a narrated video with rich explanatory voiceovers. 

For criterion (1), we employed keyframe extraction with a predefined threshold to ensure a minimum level of visual change required for keyframe selection. For newly acquired videos, we extracted keyframes using FFmpeg by computing inter-frame color histogram differences. The threshold was determined via linear interpolation between 0.008 for 5-minute videos and 0.25 for 200-minute videos. We then trained and applied a DenseNet121 image classifier to identify keyframes containing dermatological images. Videos where more than 50\% of keyframes were classified as containing dermatological content were labeled as valid.

For criterion (2), we utilized inaSpeechSegmenter to estimate the proportion of human speech within each video, setting a threshold of 0.2. Videos falling below this threshold were marked as silent or lacking sufficient explanatory narration.

\begin{figure}[t]
  \centering
  \includegraphics[width=\linewidth]{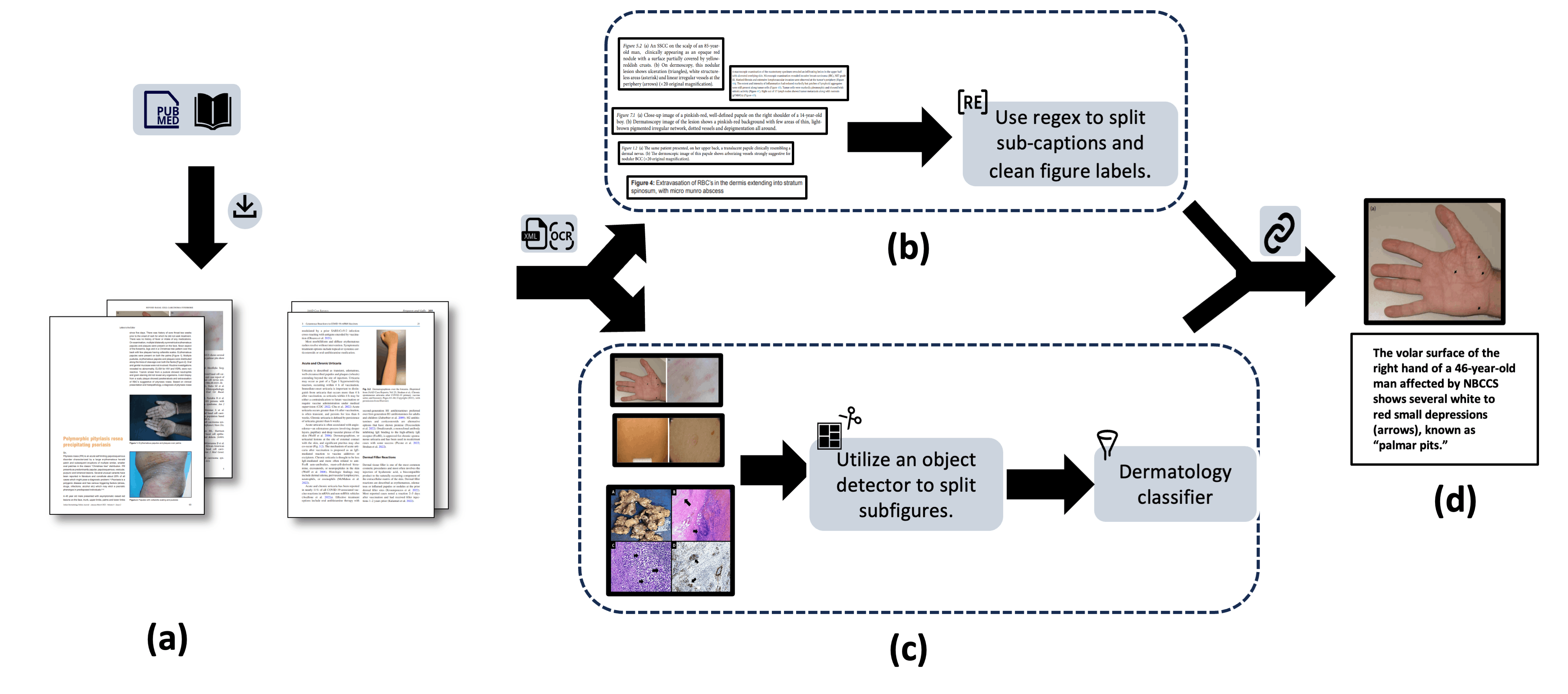}
  \caption{\textbf{Curation Pipeline for PubMed Source.} The process involves: (a) Searching and downloading articles from PubMed Open Access using pre-defined dermatology terms; (b) Extracting, splitting, and cleaning image captions; (c) Splitting subfigures using object detection and classifying skin images; and (d) Aligning image-text pairs to create the final dataset.}
  \label{fig_pubmed}
\end{figure}

\noindent \textbf{Text Extraction and Denoising.}
To address the challenges of automatic speech recognition (ASR) for medical terminology in YouTube subtitles \cite{quilt1m}, we employed the large-scale open-source Whisper Large-V3 model~\cite{radford2023robust} to perform speech-to-text conversion by directly transcribing entire audio segments. We then developed a transcription denoising and quality control pipeline consisting of three key steps:

(i) Applying the RAKE algorithm to identify key phrases (up to four words) and optimizing them by removing stop words using NLTK;

(ii) Utilizing GPT-4o to verify and correct each entry, correcting transcription errors, and refining the alignment of complete descriptive statements;

(iii) Prompting a language model to generate a structured summary of the subtitles for improved readability and organization.

\noindent \textbf{Aligning Image and Text Pairs.}
To achieve precise alignment between images and their corresponding text, we defined the timestamp interval between consecutive keyframes as an image chunk and treated each transcribed sentence as a text chunk. When the temporal overlap between an image chunk and a text chunk exceeded 50\%, we considered them a matched pair. Note that one image chunk may map to multiple text chunks, in which case we concatenated these text chunks into a single longer description.

For image chunks without any matching text chunks (i.e., those with zero temporal overlaps), we observed they typically depicted content similar to previous frames that already had mapped text (e.g., continuous discussion of the same type of lesion). Consequently, we aligned such unmatched chunks to the most recent preceding keyframe with an associated text description. To mitigate potential mismatches in fine-grained details, we additionally filtered out image-specific details from the textual content, preserving only high-level descriptive information. 

\subsection{PubMed}
Following \cite{monet,pmc-clip}, the main curation pipeline for PubMed OA is illustrated in Fig.~\ref{fig_pubmed} and detailed below.

\noindent\textbf{Collecting Image-text Pairs.} 
We retrieved dermatology-related articles published between 1990 and 2024 from the PMC Open Access Subset using 356 domain-specific terms. This query yielded 566,571 articles with approximately 3.6 million images. To filter out non-dermatology-related figures (e.g., diagrams, flow charts, cartoon illustrations, and X-rays), we implemented a combination of clustering and manual inspection. After filtering, we matched the selected images with their captions from the provided XML format files to construct approximately 50K dermatology-focused image-text pairs.

\begin{figure*}[t]
  \centering
  \includegraphics[width=0.9\linewidth]{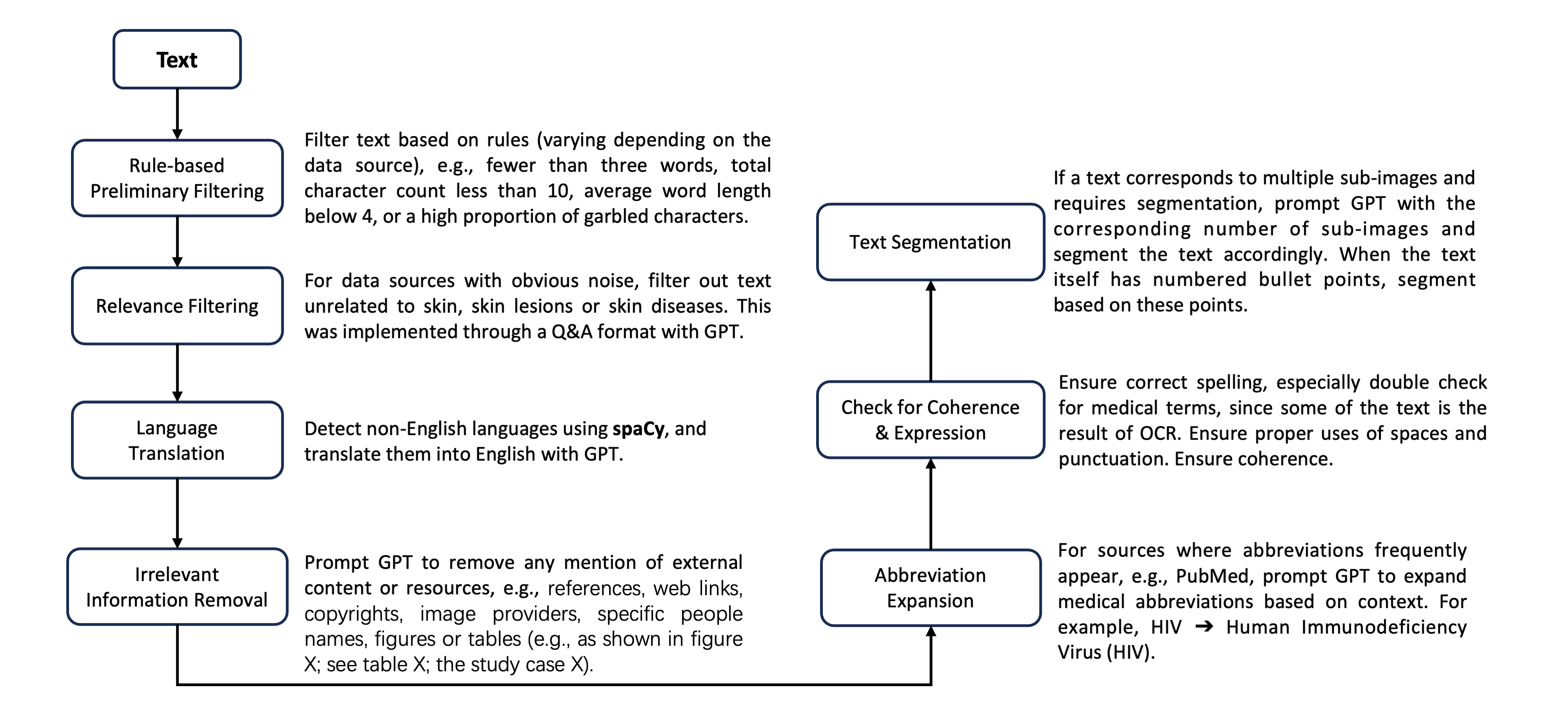}
  \caption{Workflow for general text processing.}
  \label{fig_text1}
\end{figure*}

\begin{figure*}[t]
  \centering
  \includegraphics[width=0.9\linewidth]{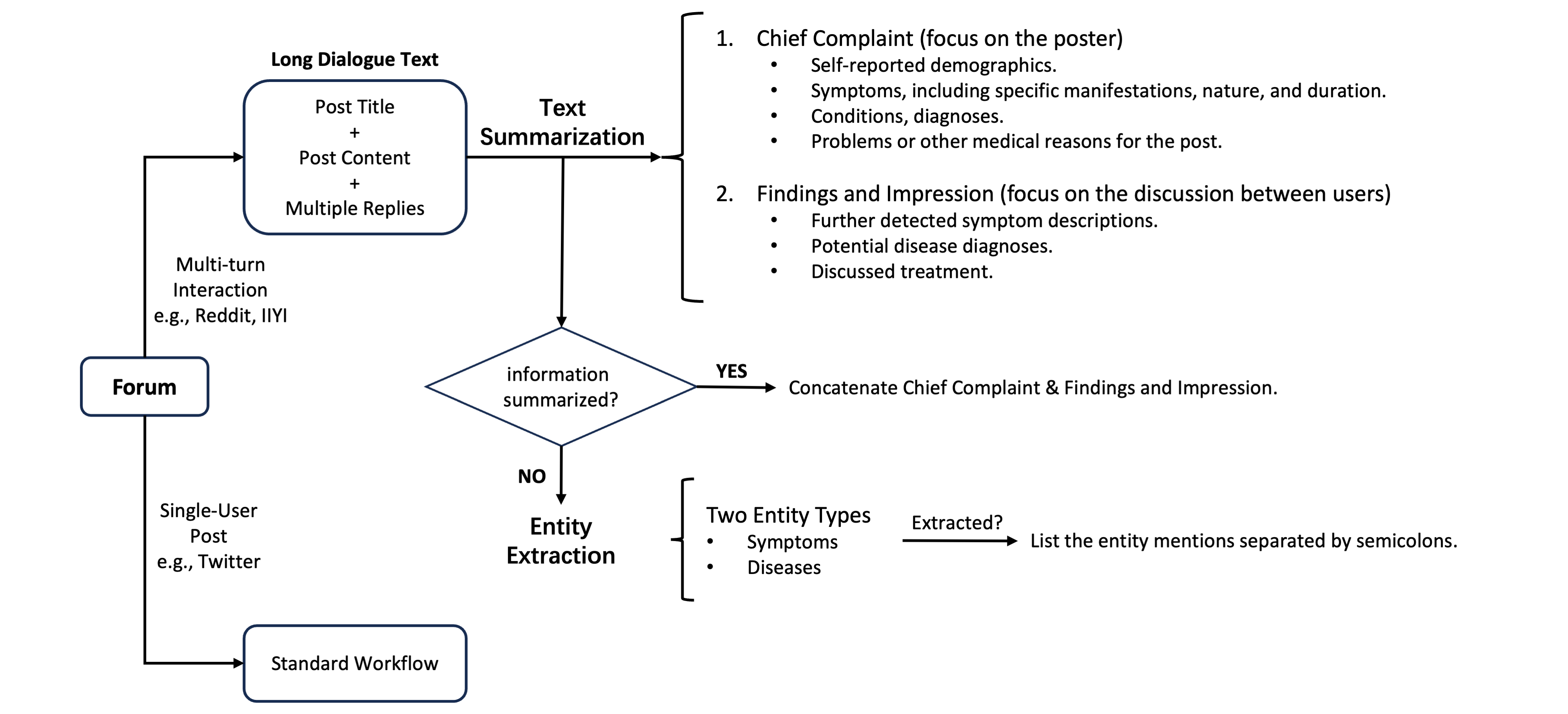}
  \caption{Workflow for forum-specific text processing.} 
  \label{fig_text2}
\end{figure*}

\noindent\textbf{Filtering Process.} 
We used EfficientNetV2-S for feature extraction and applied PCA to reduce feature dimensions to 50. Using these features, we performed hierarchical K-means clustering, first grouping images into 20 major clusters, each further divided into 20 subclusters. We manually inspected 50 representative images per subcluster, iteratively removing non-dermatology clusters over five rounds until only dermatology-related clusters remained.

\begin{figure*}[t]
  \centering
  \includegraphics[width=0.9\linewidth]{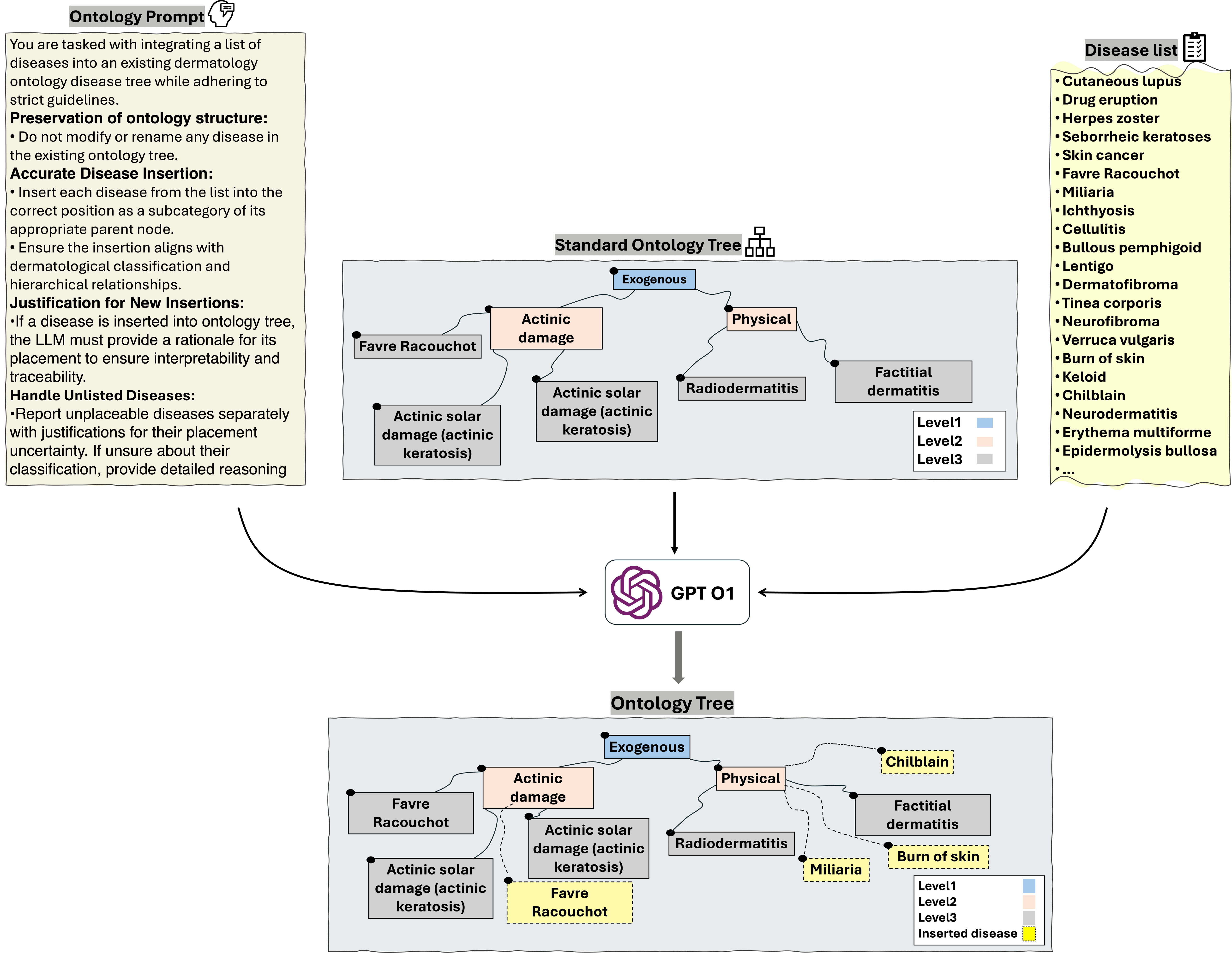}
  \caption{\textbf{Example of ontology tree construction.} A pipeline for developing a comprehensive ontology tree from a standard expert-constructed ontology tree. The LLM is provided with the standard ontology tree pre-defined by medical experts, the ontology prompt, and the standardized disease list, ensuring the original ontology structure is maintained while accurately inserting diseases into the correct hierarchical positions.}
  \label{fig_ontology}
\end{figure*}

\subsection{Educational Material}
\noindent \textbf{Collecting Image-text Pairs from Educational Material.} 
We curated image-text pairs from 68 materials using the Fitz optical scanning module from the PyMuPDF Python package. For each detected image, the nearest caption box containing figure-related patterns was automatically retrieved to form an image-text pair. When direct extraction from scanned PDFs was not feasible, Optical Character Recognition (OCR) converted them into a vectorized format before extraction.

\noindent \textbf{Removing Non-Dermatology Images.} 
To remove non-dermatology images from curated image-text pairs, we partitioned the curated image-text pairs into 20,000 image chunks for computational efficiency. EfficientNetV2-S served as a feature extractor, encoding image features that were subsequently reduced to a 50-dimensional space using Principal Component Analysis (PCA). Manual inspection and iterative non-dermatology cluster removal were performed three times to ensure the elimination of the most irrelevant images.

\noindent \textbf{Subfigure Detection and Segmentation.} 
For subfigure detection, we trained a DINO \cite{zhang2022dino} object detector using the MMDetection framework on 1,072 training images and 213 validation images sampled for each data source. The trained model then detected all subfigures, systematically cropping and arranging them in a structured left-to-right, top-to-bottom order.

\noindent \textbf{Subcaption Detection and Image-text Pairing.} 
Subcaptions were extracted using regular expressions that identified common subfigure markers (e.g., A) and (a)), facilitating automated detection and segmentation. Each subfigure was matched sequentially with its corresponding subcaption. If discrepancies arose between the number of subfigures and subcaptions, the original images and captions remained intact to preserve data integrity.

\noindent \textbf{Automated Filtering of Non-Dermatology Subfigures.} 
To further remove non-dermatology subfigures, we trained a DenseNet-121 classifier on a manually annotated dataset of 2,200 images sourced from educational materials (2,000 dermatology, 200 non-dermatology). Using a weighted random sampler and the Adam optimizer, we trained with a batch size of 128 and a learning rate of 9e-3. Early stopping was applied, halting training if validation AUROC showed no improvement after 22 epochs. By applying this classifier, we effectively excluded non-dermatology images from our dataset, ultimately ensuring a high-quality collection of image-text pairs sourced from educational materials.

\subsection{Medical Forums}
\noindent \textbf{Extracting Image-Text Pairs from Twitter Posts.} 
We began by manually reviewing content associated with 58 dermatology-related keywords to identify highly relevant channels. Through this process, we curated 27 dermatology channels comprising 14,099 posts, including both the tweet content and the three most-liked replies under each tweet. To ensure the dataset contained high-quality dermatology images, we applied the classifier mentioned in the educational materials-curated pipeline, refining the dataset to 6,726 images.

\noindent \textbf{Text Cleaning and Processing.} 
The accompanying text underwent extensive cleaning, removing @usernames, hashtags (`\#'), newline (`$\backslash n$') and carriage return (`$\backslash r$') symbols, HTML links, URLs, bold and italicized text, and other invalid characters. Additionally, all sentences containing question marks or beginning with ``What is" were eliminated to enhance textual clarity.

\noindent \textbf{Manual Removal of Advertisements.} 
Further refinement was carried out through manual inspection, leading to the removal of advertisement-related tweets, reducing the dataset to 6,532 image-text pairs. 

\noindent \textbf{Text Standardization.} 
To standardize the text, we reconstructed it by concatenating the original tweet content with its longest reply. Finally, image-text pairs containing fewer than three words were discarded, resulting in a final dataset of 6,431 high-quality image-text pairs. For other medical forums such as IIYI and Reddit, we followed similar workflows.

\subsection{Public Dataset}
Additionally, we created handcrafted image-text pairs using the publicly available SCIN \cite{scin} and MSKCC \cite{isic} datasets. For the MSKCC dataset, we generated text descriptions by integrating anatomic site, lesion type, and diagnosis results into a structured template, yielding 10,619 image-text pairs. Similarly, for the SCIN dataset, we constructed text descriptions by incorporating image modality, skin tone, age, gender, skin texture, symptoms, and diagnosis into a handcrafted template, resulting in 6,518 image-text pairs.

\subsection{Initial Text Processing and Quality Control}

\noindent\textbf{General Processing.} 
As shown in Fig.~\ref{fig_text1}, we processed non-forum text through language detection, information block filtering, and abbreviation expansion. Non-English text was identified using SpaCy and translated into English using GPT. For data sourced from PubMed and educational materials, we instructed GPT to filter out non-medical information blocks, such as citations, figure references, hyperlinks, copyright statements, and personal names. For knowledge-dense texts, GPT recognized and expanded abbreviations contextually (e.g., HIV → Human Immunodeficiency Virus (HIV)).

\begin{figure}[!t]
  \centering
  \includegraphics[width=0.9\linewidth]{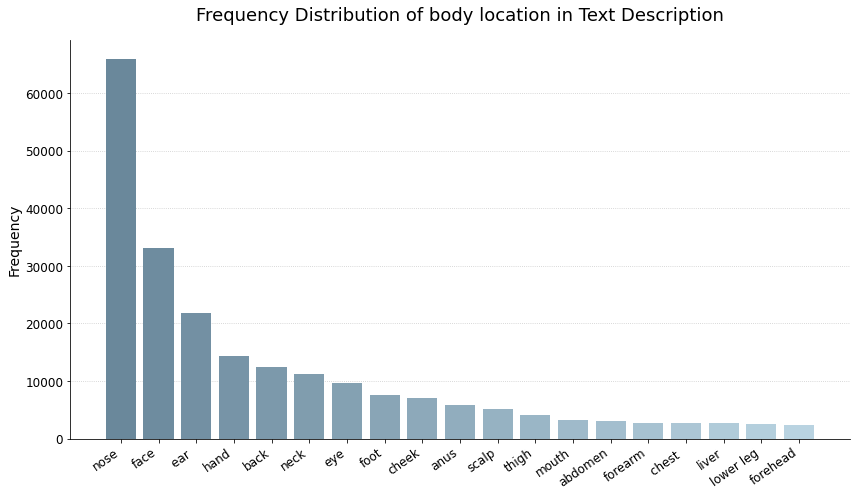}
  \caption{Frequency distribution of body sites.}
  \label{fig:loc}
\end{figure}

\begin{figure}[!t]
  \centering
  \includegraphics[width=0.9\linewidth]{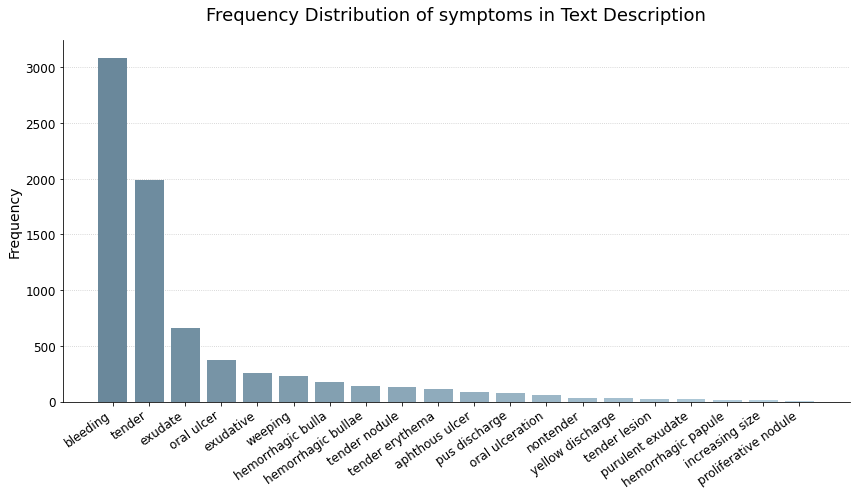}
  \caption{Frequency distribution of symptoms.}
  \label{fig:symptoms}
\end{figure}

\begin{figure*}[!t]
  \centering
  \begin{tabular}{ccc}
    \includegraphics[width=0.31\textwidth]{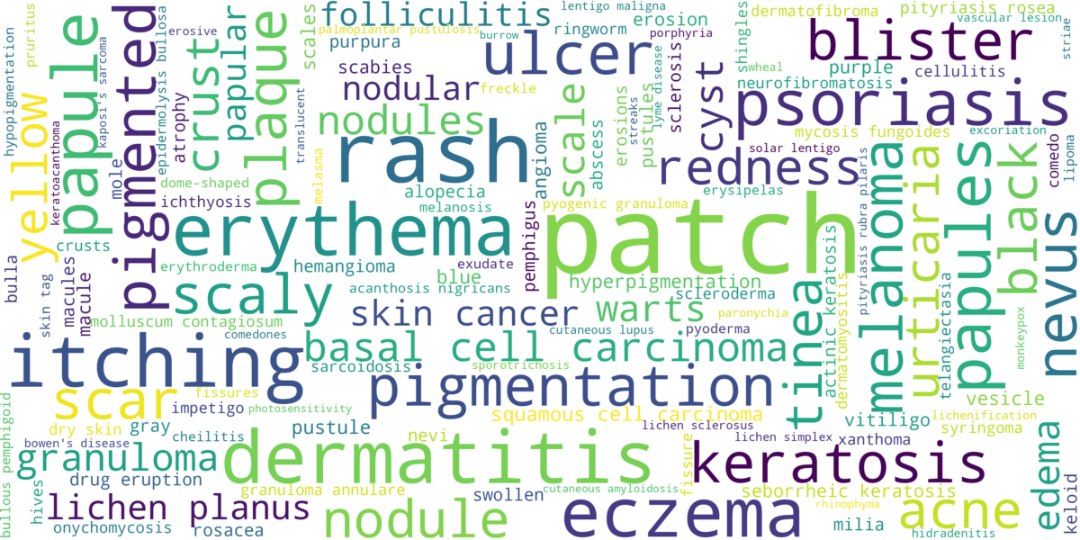} &
    \includegraphics[width=0.31\textwidth]{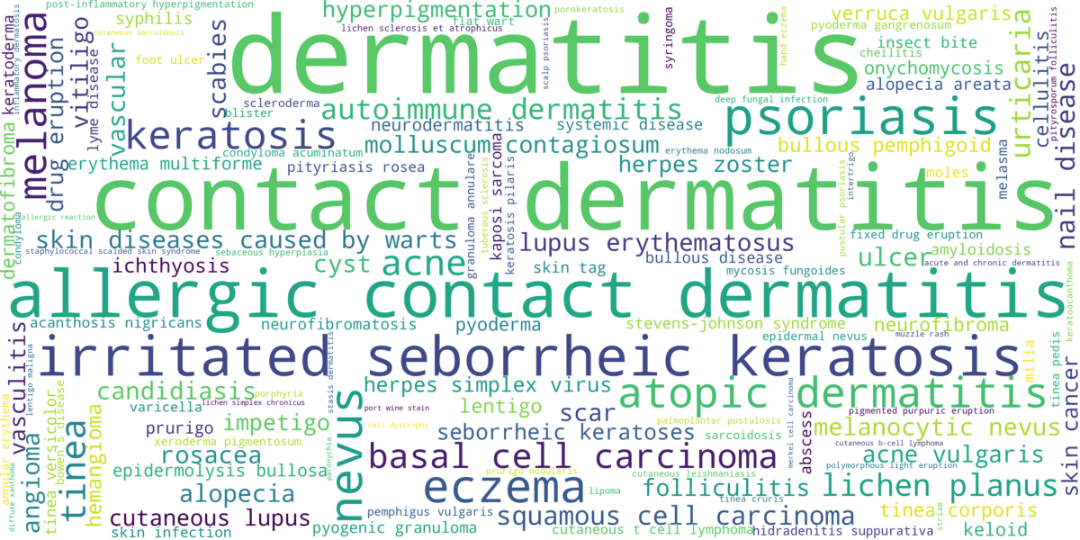} &
    \includegraphics[width=0.31\textwidth]{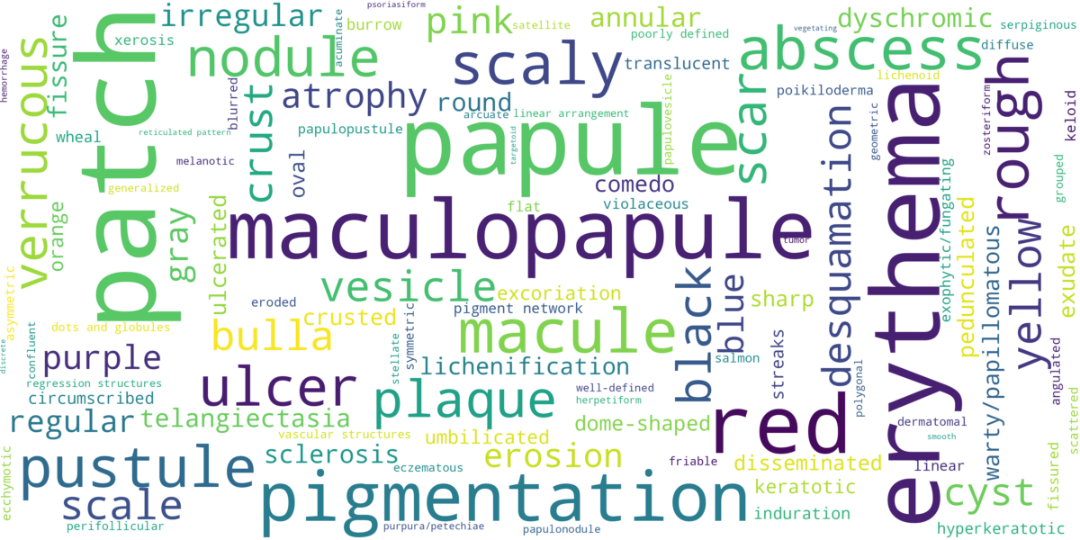} \\
  \end{tabular}
  \vspace{-2mm}
  \caption{Word cloud of medical terms: medical term (left), diseases (middle), and clinical concept (right).}
  \label{fig:three_clouds}
\end{figure*}

\noindent\textbf{Forum-Specific Processing.} 
As illustrated in Fig.~\ref{fig_text2}, for multi-turn discussions in medical forums, we extracted structured summaries, including: (1) the poster's chief complaint (demographics, symptoms, diagnosis, and medical intent) and (2) clinical findings and impressions from replies (symptom elaborations, differential diagnostic discussions, and treatment suggestions). These components were concatenated to form the final captions. If summarization failed, we instead extracted symptom and disease entities, connecting them with semicolons.

\noindent\textbf{Quality Control.} 
We applied rule-based filtering specific to data sources, including but not limited to discarding captions with fewer than three words or ten characters, as well as those consisting solely of garbled text. For noisy sources, we instructed GPT to evaluate dermatological relevance using a QA-based approach. OCR-derived text underwent spell-checking, punctuation correction, and coherence refinement. This ensured captions were precise, standardized, and dermatology-focused.

\subsection{Ontology Knowledge Augmentation}

\subsubsection{Standardized Disease and Clinical Concept}
For all data sources, the LLM-extracted content was often noisy and unformatted. We standardized diseases and concepts to avoid medical term ambiguity issues.

\noindent \textbf{Construction of the Standardized Disease List.} 
We constructed a standardized disease list by compiling disease labels from the F17k, SD128, SNU134, SCIN, and HAM datasets. Additionally, we leveraged LLM to automatically identify and merge diseases with identical content but different names, ensuring consistency and reducing redundancy. This process resulted in a standardized disease list containing 407 unique disease labels.

\noindent \textbf{Construction of Clinical Concept List.} 
We established a standardized concept list by compiling labels from public skin condition datasets, including Derm7pt and SkinCon. To further expand this list, we prompted the LLM to generate additional skin-related visual concepts based on key dermatological categories: Basic Morphology, Secondary Changes, Basic Colors, Color Characteristics, Shape Characteristics, Surface Features, Distribution Patterns, Border Characteristics, and Special Morphology. We conducted this process multiple times and manually removed any unrelated concepts. This resulted in a standardized clinical concept list containing 130 unique clinical concept labels as shown in Table~\ref{tab:concept}.

\noindent \textbf{Alignment between LLM-extracted contents and standardized lists.} 
To align LLM-extracted content with standardized lists, we implemented two distinct pipelines for disease and clinical concepts:

\textbf{1) Standardized Disease List Alignment:} We constructed a mapping framework using a Word2Vec-based approach, employing BioMedBERT as a word encoder to transform LLM-generated disease names and the standardized disease list into vector representations. To ensure accurate mappings, we iterated through the LLM-extracted content list, computing similarity scores against the standardized disease list. If the highest similarity score exceeded 0.7, the LLM-generated disease name was mapped to its corresponding standardized term. This process successfully aligned LLM-generated content with 390 unique standardized diseases as shown in Table~\ref{tab:concept}, bridging the connection between downstream classification datasets and pretrained image-text pairs.

\textbf{2) Standardized Clinical Concept List Alignment:} We used two methods to match LLM-extracted concepts with the standardized concept list. First, the Substring Matching Algorithm identified overlapping terms, successfully aligning most LLM-extracted concepts with standardized clinical concepts (e.g., ``erythematous-violaceous macule" mapped to ``erythematous," ``violaceous," and ``macule"). Second, for the remaining unmatched concepts, we employed LLM-assisted alignment, providing the LLM with both lists to iteratively refine matches through multi-turn dialogues. This process enabled the alignment of LLM-extracted concepts with 130 standardized clinical concepts.

\subsubsection{Ontology Construction and Augmentation}

To construct a dermatology ontology tree, we built upon an initial standard ontology tree (Fig.1e) curated by four dermatology experts, encompassing 128 dermatological diseases from the SD128\cite{sd198} dataset. We then utilized a specialized ontology prompt strategy that enabled the LLM to systematically integrate diseases from the standardized disease list into the ontology structure while maintaining hierarchical integrity.

\noindent \textbf{The ontology construction follows four key principles:} 
1) \textit{Preservation of the standard ontology structure} – The LLM must retain the original hierarchy and avoid modifying the positions of existing nodes. 
2) \textit{Accurate disease insertion} – Each disease from the standardized disease list must be correctly placed, considering its hierarchical relationship with existing nodes in the ontology tree. 
3) \textit{Justification for new insertions} – If a disease is inserted into the ontology tree, the LLM must provide a rationale for its placement to ensure interpretability and traceability. 
4) \textit{Handling uncertain classifications} – If the LLM is unsure of a disease's placement, it defers the decision by adding it to a separate list with an accompanying explanation.

\noindent \textbf{LLM-driven Ontology Integration.} 
As shown in Fig.~\ref{fig_ontology}, we provided the LLM with three key inputs: the standard ontology tree, the ontology prompt, and the standardized disease list. The LLM then automatically integrated diseases from the standardized disease list into the ontology tree, generating a refined structure that captured rich and diverse hierarchical relationships. For instance, Miliaria was correctly inserted as a child node under Physical and Exogenous conditions. To ensure stability and consistency, we repeated this automatic LLM-driven integration for five iterations, refining the ontology tree through iterative manual adjustments and validation. As a result, we successfully constructed an ontology tree comprising 371 skin disease conditions, while 19 general diseases remained unplaced due to the LLM's uncertainty regarding their classification. This structured methodology ensured that ontology tree development remained systematic, transparent, and aligned with expert-defined standards, while effectively leveraging the LLM's capabilities for hierarchical reasoning and disease classification.

\noindent \textbf{Ontology Caption Construction.} 
Once the standardized disease list was integrated, we used the augmented ontology tree to retrieve all parent nodes of each disease, generating hierarchical disease paths (e.g., folliculitis: inflammatory $\xrightarrow{}$ infectious $\xrightarrow{}$ bacterial $\xrightarrow{}$ folliculitis). We then transformed the hierarchy into ontology-augmented captions using a series of predefined templates, such as `This is a skin photo diagnosed as \{inflammatory, infectious, bacterial, folliculitis\}.' This approach ensured that ontology captions accurately represented hierarchical relationships within the ontology tree, providing a structured and standardized description of dermatological conditions.

\noindent \textbf{Knowledge Augmentation Caption Construction.} 
Finally, the knowledge-augmented caption was constructed by appending the ontology caption and clinical concept caption to the end of the original caption. Similar to ontology caption construction, the clinical concept captions were generated using a handcrafted template: ``This is a skin photo showing \{concept\_a, concept\_b, concept\_c\}."

\begin{figure}[!t]
  \centering
   \includegraphics[width=1\linewidth]{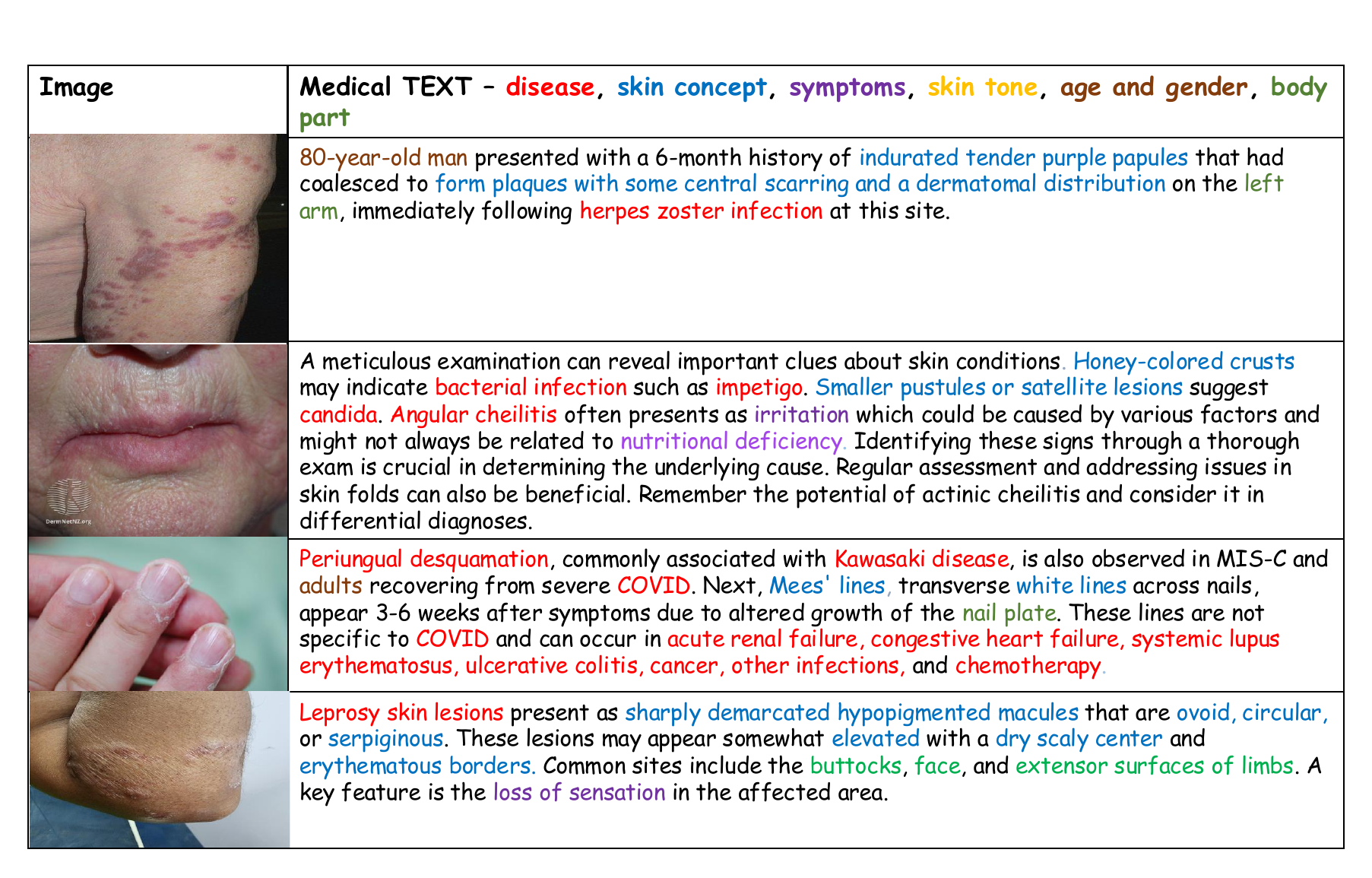}
\vspace{-7mm}
   \caption{Examples of image-text pairs from medical forums.}
   \label{supp_forum}
\end{figure}

\begin{figure}[!t]
  \centering
   \includegraphics[width=1\linewidth]{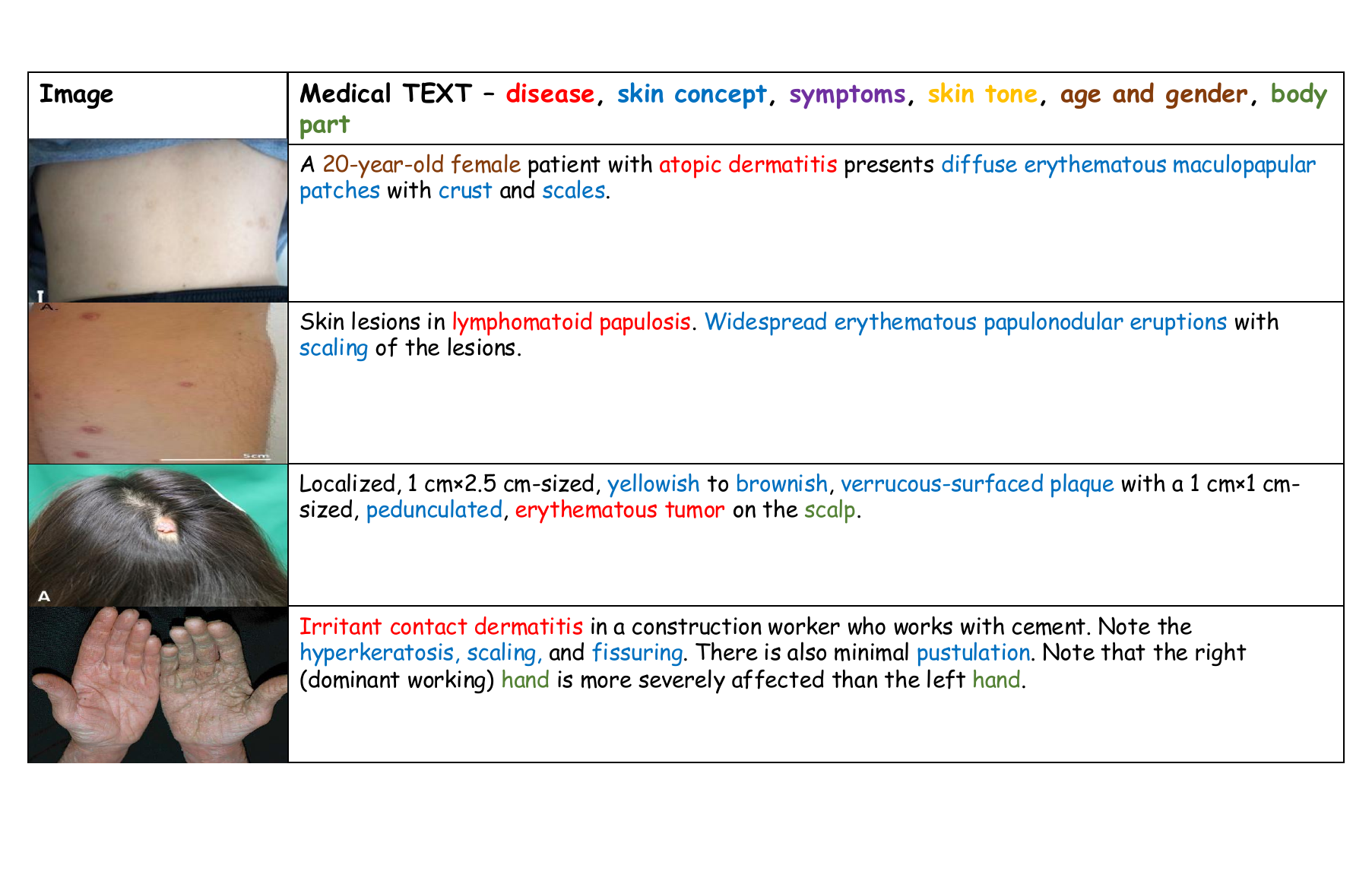}
\vspace{-13mm}
   \caption{Examples of image-text pairs from Pubmed and educational materials.}
   \label{supp_pubmed}
\end{figure}

\begin{figure}[!t]
  \centering
   \includegraphics[width=\linewidth]{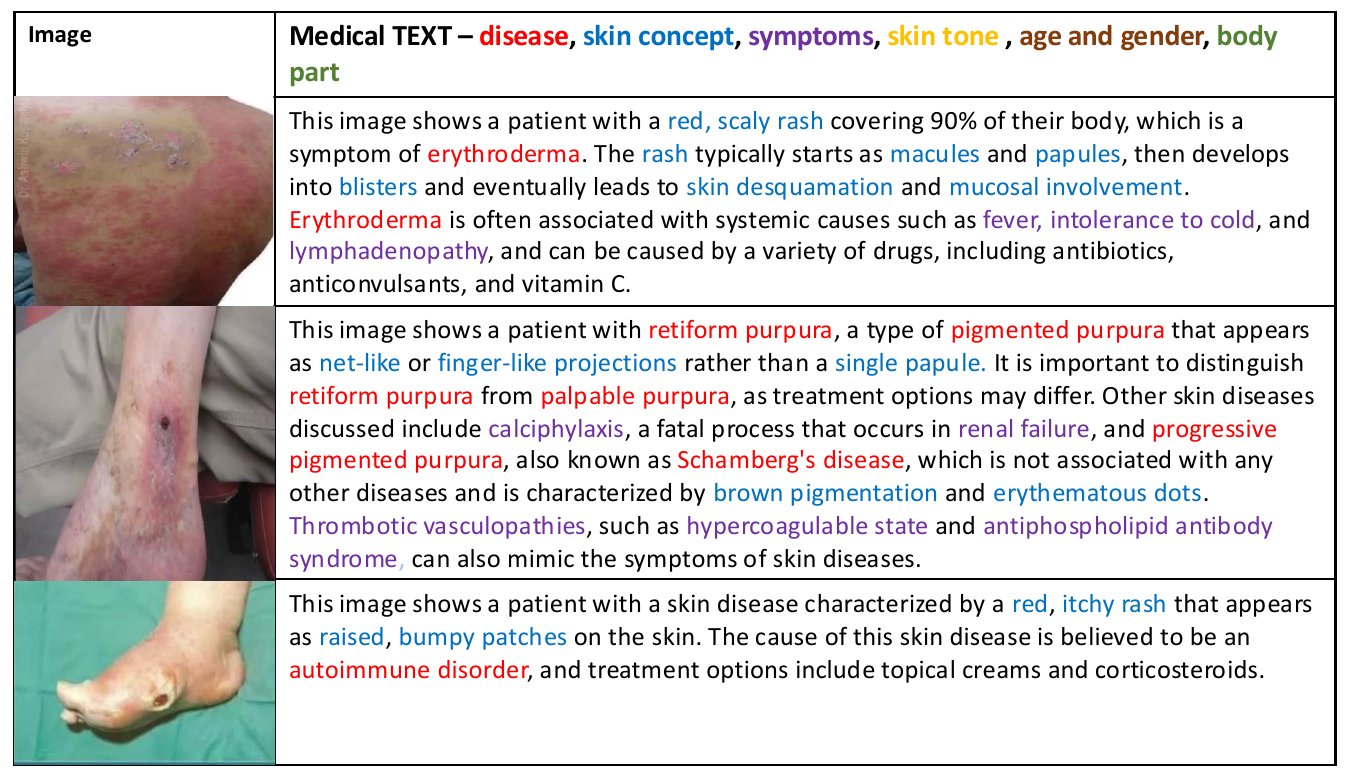}
\vspace{-6mm} \caption{Examples of image-text pairs from Youtube.}
   \label{supp_youtube}
\end{figure}

\begin{figure}[!t]
  \centering
   \includegraphics[width=1\linewidth]{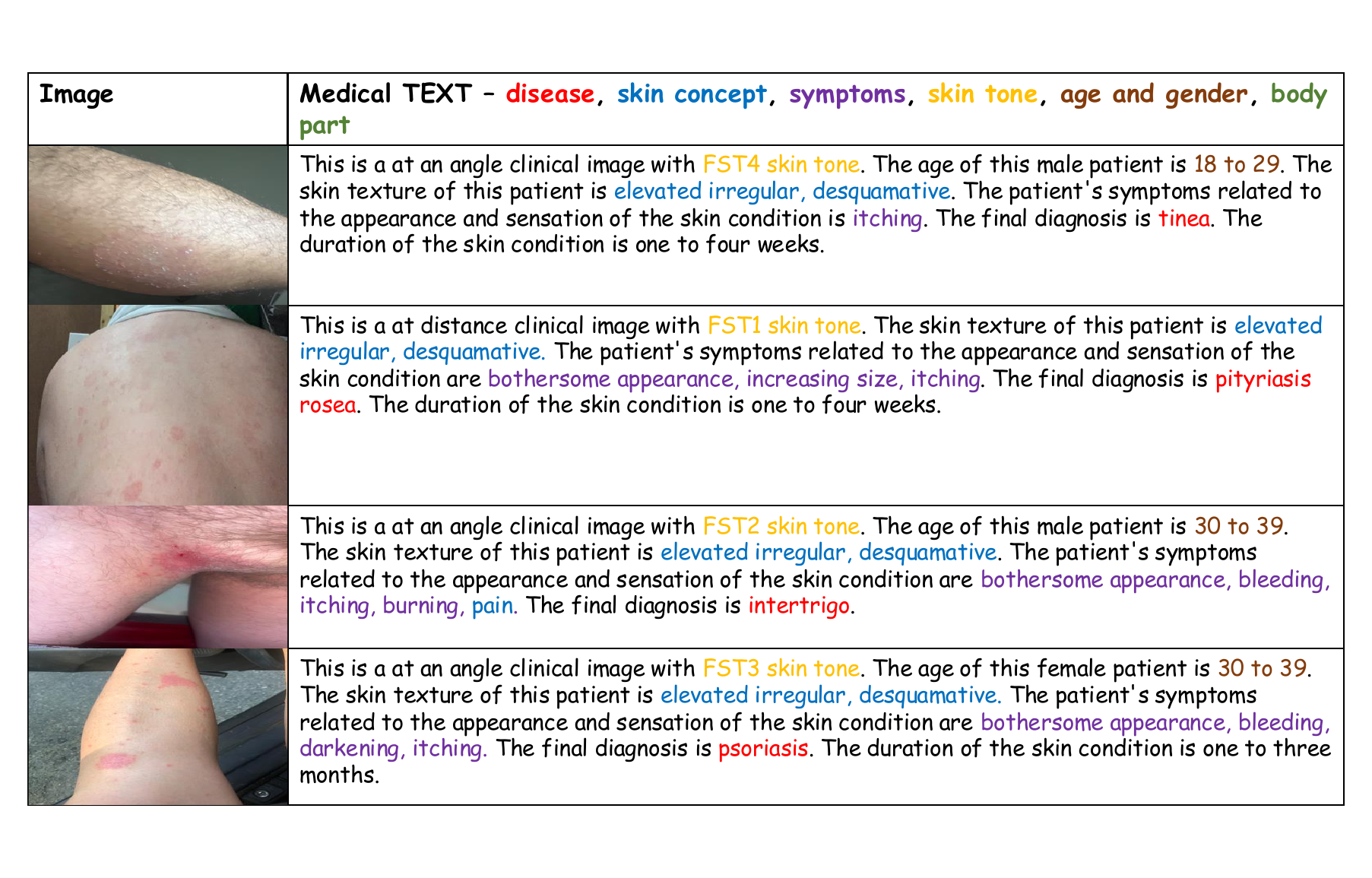}
\vspace{-11mm}
   \caption{Examples of image-text pairs re-captioned from public datasets.}
   \label{supp_public}
\end{figure}
\section{Additional Dataset Statistics}
Fig.~\ref{fig:loc} and~\ref{fig:symptoms} illustrate the frequency distribution of anatomical locations and symptoms in Derm1M. The analysis reveals that skin conditions predominantly manifest on the face, nose, and ears, while common symptoms include bleeding and tenderness. These distributions offer valuable insights into prevalence patterns within the dataset. Additionally, Fig.~\ref{fig:three_clouds} displays word clouds highlighting frequent terms across three categories: medical terminology, dermatological conditions, and clinical concepts. Fig.~\ref{supp_youtube}--\ref{supp_public} showcase representative image-text pairs from the Derm1M dataset. Table~\ref{tab:concept} and \ref{tab:disease}  show the complete list of the 390 skin conditions and 130 clinical concepts covered in Derm1M.

\section{Downstream Dataset Details}
\noindent\textbf{Daffodil}: This dataset is distinguished by its comprehensive collection of 9,548 dermatoscopic images across five skin conditions (acne, vitiligo, hyperpigmentation, nail psoriasis, and SJS-TEN), offering a valuable resource for non-melanoma skin disease classification that complements existing skin cancer-focused datasets like ISIC2019 \cite{isic} and HAM10000 \cite{ham10000}.

\section{Additional Ablation Studies}
We explore the performance differences between training methods on the Derm1M dataset, comparing SigLIP \cite{siglip}, CoCa \cite{yu2022coca}, and CLIP \cite{clip}. Tables~\ref{supp_tab1}--\ref{supp_tab3} present downstream performance across various tasks. CLIP consistently outperforms on zero-shot disease classification and few-shot/full-shot linear evaluation, achieving the highest accuracy in most settings. However, SigLIP and CoCa demonstrate superior performance on cross-modal retrieval tasks.

\section{Additional Implementation Details}

\noindent\textbf{Training Details.}
We pretrain a series of models called DermLIP on the Derm1M dataset following CLIP \cite{clip}'s contrastive learning objective. Each model is trained for 30 epochs on a single NVIDIA H200 GPU. We swap hyperparameters including batch size and learning rate, selecting the best-performing models based on validation loss.

\noindent\textbf{Prompt Details for Zero-shot Classification}
We adhere to the zero-shot classification method of the OpenCLIP framework, utilizing a prompt ensemble strategy for evaluation. The specific prompt templates employed in this process are detailed in Table~\ref{tab:prompt-templates}.

\noindent\textbf{Hyper-parameter tables for main models}
We present the pre-training hyper-parameters for the DermLIP models in Table~\ref{tab:hyperparameters_combined}. The table includes all critical training hyper-parameters, while the remaining parameters adhere to the default settings of the OpenCLIP framework.

\begin{table*}[t]
\footnotesize
\centering
\setlength{\tabcolsep}{6pt}
% \begin{tabular}{p{1.6cm}|p{1.7cm}|p{1.7cm}|p{1.7cm}|ccccc}
\begin{tabular}{p{2.3cm}|l|l|l|ccccc}
\hline
Training methods & Pretrained Data & Vision Enc. & Text Enc. & HAM & F17K & PAD & Daffodil & Average \\
\hline

\#class&   &  & & 7 & 113 &6 & 5 & \\
\hline
SigLIP &  Derm1M & ViT-B16 & SigLIP & 0.6068 & 0.2249 & 0.5857 & 0.7058 & 0.5308 \\
CoCa & Derm1M & ViT-B32 & GPT77 & 0.4098 & 0.1700 & 0.5466 & \textbf{0.7262} & 0.4632 \\ 
CLIP &Derm1M& ViT-B16 & GPT77 & \textbf{0.6820} & \textbf{0.2278} & \textbf{0.6074} & 0.7257 & \textbf{0.5607} \\
\hline
\end{tabular}
\caption{Ablation on different training methods for zero-shot disease classification (Acc).}
\label{supp_tab1}
\end{table*}

\begin{table*}[t]
\footnotesize
\centering
\setlength{\tabcolsep}{6pt}
\begin{tabular}{c|l|l|l|llllll}
\hline
\makecell{Labeling\\Ratio} & Methods & Vision Enc. & Text Enc. & HAM & F17K & PAD & Daffodil & Average \\
\hline
\#class &   &  & & 7 & 113 &6 & 5 & \\
\hline
\multirow{4}{*}{1\%} 
& SigLIP & ViT-B16 & SigLIP & 0.6986 & 0.1394 & 0.5098 & 0.7476 & 0.5239 \\
& CoCa & ViT-B32 & GPT77 & 0.7212 & 0.1349 & 0.5076 & 0.7974 & 0.5403 \\
& CLIP & ViT-B16 & GPT77 & \textbf{0.7458} & \textbf{0.1602} & \textbf{0.5184} & \textbf{0.8545} & \textbf{0.5697} \\
\hline
\multirow{4}{*}{10\%}
& SigLIP & ViT-B16 & SigLIP & 0.8037 & 0.2980 & 0.6312 & 0.8759 & 0.6522 \\
& CoCa & ViT-B32 & GPT77 & 0.7532 & 0.2967 & 0.6551 & 0.8681 & 0.6433 \\
& CLIP & ViT-B16 & GPT77 & \textbf{0.8110} & \textbf{0.3555} & \textbf{0.6594} & \textbf{0.9372} & \textbf{0.6908} \\
\hline
\multirow{4}{*}{100\%}
& SigLIP & ViT-B16 & SigLIP & \textbf{0.8550} & 0.4433 & 0.6703 & 0.9330 & 0.7254 \\
& CoCa & ViT-B32 & GPT77 & 0.7591 & 0.4933 & 0.7115 & 0.8743 & 0.7096 \\
& CLIP & ViT-B16 & GPT77 & 0.8523 & \textbf{0.5102} & \textbf{0.7614} & \textbf{0.9644} & \textbf{0.7720} \\
\hline
\end{tabular}
\caption{Ablation on different training methods for linear evaluation (Acc).}
\label{supp_tab2}
\end{table*}

\begin{table*}[t]
\footnotesize
\centering
\setlength{\tabcolsep}{6pt}
\begin{tabular}{l|l|l|cccc|cccc}
\hline
\multirow{3}{*}{Training methods} & \multirow{3}{*}{Vision Enc.} & \multirow{3}{*}{Text Enc.} & \multicolumn{4}{c|}{Holdout (n=9806)} & \multicolumn{4}{c}{SkinCAP (n=3989)} \\
\cline{4-11}
& & & \multicolumn{2}{c}{I2T (\%)} & \multicolumn{2}{c|}{T2I (\%)} & \multicolumn{2}{c}{I2T (\%)} & \multicolumn{2}{c}{T2I (\%)} \\
\cline{4-11}
& & & R@10 & R@50 & R@10 & R@50 & R@10 & R@50 & R@10 & R@50 \\
\hline
SigLIP & ViT-b16 & SigLIP & 0.3763 & 0.5614 & 0.3818 & 0.5716 & \textbf{0.1860} & \textbf{0.3896} & \textbf{0.1908} & \textbf{0.4141} \\
CoCa & ViT-B32 & GPT77 & \textbf{0.4150} & \textbf{0.6102} & \textbf{0.4182} & \textbf{0.6116} & 0.1564 & 0.3643 & 0.1737 & 0.3818 \\
CLIP & ViT-b16 & GPT77 & 0.4069 & 0.6021 & 0.3966 & 0.5992& 0.1567 & 0.3632 & 0.1594 & 0.3567 \\
\hline
\end{tabular}
\caption{Ablation on different training methods for cross-modal retrieval results. I2T represents image-to-text retrieval and T2I represents text-to-image retrieval.}
\label{supp_tab3}
\end{table*}

\begin{table*}[t]
\footnotesize
\centering
\setlength{\tabcolsep}{2pt}
\begin{tabular}{l|cc|cc|c}
\hline
\multirow{3}{*}[-1ex]{\centering\textbf{Method}} & \multicolumn{2}{c|}{\textbf{Encoder}} & \multicolumn{3}{c}{\textbf{AUROC}} \\
\cline{2-6}
& \multirow{2}{*}[-0.5ex]{\centering\textbf{Vision}}
  & \multirow{2}{*}[-0.5ex]{\centering\textbf{Text}}
  & \textbf{SkinCon}
  & \textbf{Derm7pt}
  & \multirow{2}{*}[-0.5ex]{\centering\textbf{Average}} \\
 & & & \textbf{(32)} & \textbf{(7)} & \\
\hline
CLIP-B16 \cite{clip} & ViT-B16 & GPT77 & 0.6643 & 0.5594 & 0.6119 \\
SigLIP \cite{siglip} & ViT-B16 & SigLIP & 0.6769 & 0.5631 & 0.6200 \\
CoCa \cite{yu2022coca} & ViT-B32 & GPT77 & 0.6041 & 0.5677 & 0.5859 \\
\hline
PMC-CLIP \cite{pmc-clip} & ResNet50 & GPT77 & 0.6251 & 0.5820 & 0.6036 \\
BiomedCLIP \cite{biomedclip} & ViT-B16 & PMB256 & 0.6817 & 0.6092 & 0.6455 \\
MONET \cite{monet} & ViT-L14 & GPT77 & 0.7502 & \textbf{0.6889} & 0.7196 \\
\hline
\cellcolor{gray!15}DermLIP & \cellcolor{gray!15}ViT-B16 & \cellcolor{gray!15}GPT77 & \cellcolor{gray!15}\textbf{0.7728} & \cellcolor{gray!15}0.6877 & \cellcolor{gray!15}\textbf{0.7303} \\
\cellcolor{gray!15}DermLIP & \cellcolor{gray!15}PanDerm-B & \cellcolor{gray!15}PMB256 & \cellcolor{gray!15}0.7299 & \cellcolor{gray!15}0.6148 & \cellcolor{gray!15}0.6724 \\
\hline
\end{tabular}
\vspace{-2mm}
\caption{Zero-shot concept annotation (AUROC).}
\label{tab:concept}
\end{table*}

% \begin{figure*}[!t]
%   \centering
%    \includegraphics[width=1\linewidth]{supp_figures/artifacts_factor_annotation.png}
%    \caption{Artificial Factors Annotation Examples: Comparison of CLIP-ViT-B/16 Models Pre-trained on the Derm1M Dataset and MONET}
%    \label{fig_artifacial_factor_annotation}
% \end{figure*}

% \begin{figure*}[t!]
%     \centering
%     \setlength{\tabcolsep}{5pt}  % Adjust this value to get desired spacing
%     \renewcommand{\arraystretch}{0.8}
%     \begin{tabular}{c@{\hspace{10pt}}c}
%         \includegraphics[width=0.476\linewidth]{supp_figures/dermlip_artifacts.png} &
%         \includegraphics[width=0.476\linewidth]{supp_figures/monet_artifacts.png} \\
%         \footnotesize{(a) DermLIP (Ours)} & \footnotesize{(b) MONET.} \\
%     \end{tabular}
% \vspace{-2mm}
% \caption{\small \textbf{Comparison of DermLIP and MONET on concept annotations.}
%     \label{fig5}
% \end{figure*}

\begin{table}[ht]
\footnotesize
\centering
\begin{tabular}{|p{4cm}|p{4cm}|p{4cm}|p{4cm}|}
\hline
\cellcolor{gray!25}\textbf{A-E} & \cellcolor{gray!25}\textbf{F-M} & \cellcolor{gray!25}\textbf{N-R} & \cellcolor{gray!25}\textbf{S-Z} \\
\hline
abscess & fissure & necrosis & salmon \\
acuminate & fissured & nodule & satellite \\
angulated & flat & orange & scale \\
annular & flat topped & oval & scaly \\
arciform arrangement & follicular-centered & papule & scar \\
arcuate & friable & papulonodule & scattered \\
asymmetric & generalized & papulopustule & sclerosis \\
atrophy & geometric & papulovesicle & serpiginous \\
black & gray & patch & sharp \\
blue & grouped & pedunculated & smooth \\
blue whitish veil & hemorrhage & perifollicular & stellate \\
blurred & herpetiform & pigment network & streaks \\
brown(hyperpigmentation) & hyperkeratotic & pigmentation & symmetric \\
bulla & induration & pink & targetoid \\
burrow & irregular & plaque & telangiectasia \\
circumscribed & keloid & poikiloderma & translucent \\
clustered & keratotic & polygonal & tumor \\
comedo & leaf-shaped & poorly defined & ulcer \\
confluent & lichenification & psoriasiform & ulcerated \\
crust & lichenoid & purple & umbilicated \\
crusted & linear & purpura/petechiae & vascular structures \\
cyst & linear arrangement & pustule & vegetating \\
dermatomal & localized & raised & verrucous \\
desquamation & macule & red & vesicle \\
diffuse & maculopapule & regression structures & violaceous \\
discrete & maculopatch & regular & warty/papillomatous \\
disseminated & melanotic & reticular & wedge-shaped \\
dome-shaped & molluscoid & reticulated pattern & well-defined \\
dots and globules &  & rough & wheal \\
dyschromic &  & round & white(hypopigmentation) \\
ecchymotic &  &  & xanthomatous \\
eczematous &  &  & xerosis \\
eroded &  &  & yellow \\
erosion &  &  & zosteriform \\
erythema &  &  &  \\
excoriation &  &  &  \\
exophytic/fungating &  &  &  \\
exudate &  &  &  \\
\hline
\end{tabular}
\caption{Full List of 130 standardized clinical concepts.}
\label{tab:concept}
\end{table}

\begin{onecolumn}
\footnotesize
\centering
\begin{longtable}{|p{7cm}|p{9.5cm}|}
\hline
\cellcolor{gray!25}\textbf{A-E} & \cellcolor{gray!25}\textbf{F-M} \\
\hline
abscess & factitial dermatitis \\
acanthosis nigricans & favre racouchot \\
acne & fibroma molle \\
acne keloidalis nuchae & fixed drug eruption \\
acne urticata & fixed eruptions \\
acne vulgaris & flat wart \\
acquired autoimmune bullous diseaseherpes gestationis & flushing \\
acrokeratosis verruciformis & follicular mucinosis \\
actinic granuloma & folliculitis \\
actinic solar damage(actinic keratosis) & foot ulcer \\
actinic solar damage(cutis rhomboidalis nuchae) & foreign body reaction of the skin \\
actinic solar damage(pigmentation) & fox-fordyce disease \\
actinic solar damage(solar elastosis) & freckle \\
actinic solar damage(solar purpura) & fungal dermatitis \\
actinic solar damage(telangiectasia) & fungal dermatosis \\
acute and chronic dermatitis & furuncle \\
acute constitutional eczema & geographic tongue \\
acute dermatitis & granulation tissue \\
acute dermatitis, nos & granuloma annulare \\
acute generalized exanthematous pustulosis & granuloma faciale \\
acute vesicular dermatitis & grover's disease \\
adnexal neoplasm & guttate psoriasis \\
ageing skin & hailey hailey disease \\
allergic contact dermatitis & halo nevus \\
allergic reaction & hand eczema \\
alopecia & hand foot and mouth disease \\
alopecia areata & hemangioma \\
alopecia mucinosa & hematoma of skin \\
amyloidosis & hemosiderin pigmentation of lower limb due to varicose veins of lower limb \\
angiofibroma & hemosiderin pigmentation of skin due to venous insufficiency \\
angiokeratoma & herpes simplex virus \\
angioma & herpes zoster \\
angular cheilitis & hidradenitis suppurativa \\
animal bite - wound & histiocytosis of skin \\
annular erythema & hormonal acne \\
apocrine hydrocystoma & hyperkeratosis palmaris et plantaris \\
arsenical keratosis & hyperpigmentation \\
atopic dermatitis & hypersensitivity \\
atopic winter feet & hypertrichosis \\
autoimmune dermatitis & hypertrophic scar \\
basal cell carcinoma & ichthyosis \\
beau's lines & idiopathic guttate hypomelanosis \\
becker nevus & impetigo \\
behcets disease & infantile atopic dermatitis \\
benign keratosis & infected eczema \\
blister & inflammatory dermatosis \\
blue nevus & insect bite \\
bowen's disease & intertrigo \\
bullous disease & inverse psoriasis \\
bullous pemphigoid & irritant contact dermatitis \\
burn of forearm & irritated seborrheic keratosis (from "sk/isk") \\
burn of skin & junction nevus \\
café au lait macule & juvenile plantar dermatosis \\
calcinosis cutis & juvenile xanthogranuloma \\
callus & kaposi sarcoma \\
campbell de morgan spots & kaposi's sarcoma of skin \\
candida intertrigo & keloid \\
candidiasis & keratoacanthoma \\
cellulitis & keratoderma \\
central centrifugal cicatricial alopecia & keratolysis exfoliativa of wende \\
cheilitis & keratosis \\
chilblain & keratosis pilaris \\
childhood bullous pemphigoid & keratosis pilaris rubra faciei \\
cholestasis of pregnancy & kerion \\
chondrodermatitis nodularis helicis & knuckle pads \\
chronic actinic dermatitis & koilonychia \\
chronic dermatitis, nos & langerhans cell histiocytosis \\
clubbing of fingers & leg veins \\
compound nevus & lentigo \\
condyloma & lentigo maligna \\
condyloma acuminatum & lentigo maligna melanoma \\
confluent and reticulated papillomatosis & leukocytoclastic vasculitis \\
congenital nevus & leukonychia \\
contact dermatitis & lichen amyloidosis \\
contact dermatitis caused by rhus diversiloba & lichen nitidus \\
contact dermatitis, nos & lichen planus \\
contact purpura & lichen sclerosis et atrophicus \\
crowe's sign & lichen simplex chronicus \\
cutaneous b-cell lymphoma & lichen spinulosus \\
cutaneous horn & lichen striatus \\
cutaneous larva migrans & lipoma \\
cutaneous leishmaniasis & livedo reticularis \\
cutaneous lupus & local infection of wound \\
cutaneous sarcoidosis & localized cutaneous vasculitis \\
cutaneous t cell lymphoma & localized skin infection \\
cyst & lupus erythematosus \\
darier-white disease & lyme disease \\
dariers disease & lymphangioma \\
deep fungal infection & lymphocytic infiltrate of jessner \\
degos disease & majocchi granuloma \\
dermatitis & median nail dystrophy \\
dermatitis herpetiformis & medication-induced cutaneous pigmentation \\
dermatofibroma & melanin pigmentation due to exogenous substance \\
dermatosis papulosa nigra & melanocytic nevus \\
desquamation & melanoma \\
diffuse xanthoma & melasma \\
digital fibroma & merkel cell carcinoma \\
dilated pore of winer & metastatic carcinoma \\
discoid eczema & milia \\
disseminated actinic porokeratosis & miliaria \\
drug eruption & moles \\
drug eruptions \& reactions & molluscum contagiosum \\
drug-induced pigmentary changes & morphea \\
dry skin & mucinosis \\
dyshidrosiform eczema & mucocele \\
dysplastic nevus & mucosal melanotic macule \\
ecthyma & muzzle rash \\
ecthyma gangrenosum & mycosis fungoides \\
eczema & myxoid cyst \\
eczema herpeticum &  \\
ehlers danlos syndrome &  \\
elephantiasis nostras &  \\
epidermal nevus &  \\
epidermoid cyst &  \\
epidermolysis bullosa &  \\
erosion of skin &  \\
erosive pustular dermatosis of the scalp &  \\
eruptive odontogenic cyst &  \\
eruptive xanthoma &  \\
erythema ab igne &  \\
erythema annulare centrifugum &  \\
erythema craquele &  \\
erythema dyschromicum perstans &  \\
erythema elevatum diutinum &  \\
erythema gyratum repens &  \\
erythema migrans &  \\
erythema multiforme &  \\
erythema nodosum &  \\
exfoliative dermatitis &  \\
exfoliative erythroderma &  \\
\hline \cellcolor{gray!25}\textbf{N-R} & \cellcolor{gray!25}\textbf{S-Z} \\
\hline
naevus comedonicus & sand-worm eruption \\
nail disease & sarcoidosis \\
nail dystrophy & scabies \\
nail psoriasis & scalp psoriasis \\
necrobiosis lipoidica & scar \\
nematode infection & scleroderma \\
neurodermatitis & scleromyxedema \\
neurofibroma & sebaceous hyperplasia \\
neurofibromatosis & seborrheic keratoses \\
neutrophilic dermatoses & sixth disease \\
nevus & skin and soft tissue atypical mycobacterial infection \\
nevus depigmentosus & skin cancer \\
nevus sebaceous of jadassohn & skin diseases caused by warts \\
nevus spilus & skin infection \\
no definitive diagnosis & skin lesion in drug addict \\
nummular eczema & skin tag \\
onycholysis & spider veins \\
onychomycosis & squamous cell carcinoma \\
onychoschizia & staphylococcal scalded skin syndrome \\
organoid nevus & stasis dermatitis \\
ota nevus & stasis edema \\
others & stasis ulcer \\
palmoplantar pustulosis & steatocystoma multiplex \\
palpable migrating erythema & steroid acne \\
papular dermatoses of pregnancy & steroid use abusemisuse dermatitis \\
parapsoriasis & stevens-johnson syndrome \\
paronychia & strawberry birthmarks \\
parvovirus b19 infection & striae \\
pemphigus vulgaris & subungual hematoma \\
phototherapy & sun spots \\
phytophotodermatitis & sunburn \\
pigmentation of pregnancy & superficial gyrate erythema \\
pigmented progressive purpuric dermatosis & superficial spreading melanoma ssm \\
pigmented purpuric eruption & superficial wound of body region \\
pilar cyst & sweet syndrome \\
pincer nail deformity & sweet’s syndrome \\
pityriasis alba & syphilis \\
pityriasis lichenoides & syringoma \\
pityriasis lichenoides chronica & systemic disease \\
pityriasis lichenoides et varioliformis acuta & telangiectasia macularis eruptiva perstans \\
pityriasis rosea & tick bite \\
pityriasis rubra pilaris & tinea \\
pityrosporum folliculitis & tinea corporis \\
poikiloderma & tinea cruris \\
poikiloderma of civatte & tinea manus \\
poisoning by nematocyst & tinea pedis \\
polymorphic eruption of pregnancy & tinea versicolor \\
polymorphous light eruption & transient acantholytic dermatosis \\
porokeratosis & traumatic blister \\
porokeratosis of mibelli & traumatic ulcer \\
poroma & tuberous sclerosis \\
porphyria & tungiasis \\
port wine stain & ulcer \\
post-inflammatory hyperpigmentation & unilateral laterothoracic exanthem \\
post-inflammatory hypopigmentation & urticaria \\
post-inflammatory pigmentation & urticaria pigmentosa \\
pressure ulcer & urticarial vasculitis \\
prurigo & varicella \\
prurigo gravidarum & varicose veins of lower extremity \\
prurigo nodularis & vascular \\
prurigo of pregnancy & vasculitis \\
prurigo pigmentosa & venous lake \\
pruritic urticarial papules and plaques of pregnancy & verruca vulgaris \\
pruritus ani & viral exanthem \\
pseudo-glucagonoma syndrome & viral exanthems: roseola \\
pseudofolliculitis barbae & vitiligo \\
pseudorhinophyma & wound/abrasion \\
psoriasis & xanthelasma \\
pustular psoriasis & xeroderma pigmentosum \\
pyoderma & xerosis \\
pyoderma gangrenosum & xerotic eczema \\
pyogenic granuloma &  \\
radiodermatitis &  \\
raynaud phenomenon &  \\
red stretch marks &  \\
relapsing polychondritis &  \\
rheumatoid nodule &  \\
rhinophyma &  \\
riehl melanosis &  \\
rosacea &  \\
\hline
\caption{Full list of 390 standardized skin conditions.}
\label{tab:disease}
\end{longtable}
\end{onecolumn}

\begin{table}[htbp]
\centering
\begin{tabular}{ll}
\toprule
\textbf{ID} & \textbf{Template} \\
\midrule
1 & \texttt{This is a skin image of \{CLASS\_LABEL\}.} \\
2 & \texttt{This is a skin image of \{CLASS\_LABEL\}.} \\
3 & \texttt{A skin image of \{CLASS\_LABEL\}.} \\
4 & \texttt{An image of \{CLASS\_LABEL\}, a skin condition.} \\
5 & \texttt{\{CLASS\_LABEL\}, a skin disorder, is shown in this image.} \\
6 & \texttt{The skin lesion depicted is \{CLASS\_LABEL\}.} \\
7 & \texttt{The skin cancer in this image is \{CLASS\_LABEL\}.} \\
8 & \texttt{This image depicts \{CLASS\_LABEL\}, a type of skin cancer.} \\
\bottomrule
\end{tabular}
\caption{Prompt templates for zero-shot classification.}
\label{tab:prompt-templates}
\end{table}

\begin{table}[ht]
\centering
\renewcommand{\arraystretch}{1.2}  % 增加行高
\begin{tabular}{ccccc}
\toprule
\multicolumn{1}{c}{\textbf{Hyper-parameters}} & 
\multicolumn{1}{c}{\textbf{ViT-B16 + GPT77}} & 
\multicolumn{1}{c}{\textbf{PanDerm-B + PMB256}} & 
\multicolumn{1}{c}{\textbf{ViT-B16 + SigLIP}} & 
\multicolumn{1}{c}{\textbf{ViT-B32 + GPT77}} \\
\midrule
warmup & 1000 & 1000 & 1000 & 1000 \\ 
weight decay & 0.1 & 0.1 & 0.1 & 0.1 \\ 
LR Scheduler & cosine & cosine & cosine & cosine \\
batch size & 4096 & 2048 & 2048 & 512 \\ 
learning rate & 1e-4 & 1e-4 & 1e-4 & 1e-4 \\ 
epochs & 30 & 30 & 30 & 30 \\ 
\midrule
Pretrain & openai & PanDerm & webli & laion2b\_s13b\_b90k \\ 
Vision Encoder & ViT-B16 & PanDerm-B & ViT-B16 & ViT-B32 \\ 
Text Encoder & GPT77 & PMB256 & SigLIP & GPT77 \\ 
\bottomrule
\end{tabular}
\caption{Hyperparameters for DermLIP models pretraining.}
\label{tab:hyperparameters_combined}
\end{table}

\end{document}